\pdfoutput=1

\documentclass[11pt]{article}
\usepackage{lmodern}
\usepackage[final]{acl}
\usepackage{times}
\usepackage{latexsym}
\usepackage{booktabs}
\usepackage{graphicx} 
\usepackage{colortbl}
\usepackage{multirow}
\usepackage{xcolor}
\usepackage{amsmath} 
\usepackage{subfigure}
\usepackage{makecell}
\usepackage{bm}
\usepackage{pifont}

\usepackage{varwidth}
\usepackage{tikz}
\usetikzlibrary{calc}
\usetikzlibrary{positioning, shapes, fit, backgrounds, decorations.pathreplacing}
\NewDocumentCommand\perologo{}{\includegraphics[scale=0.028,trim=0 150 0 50,clip]{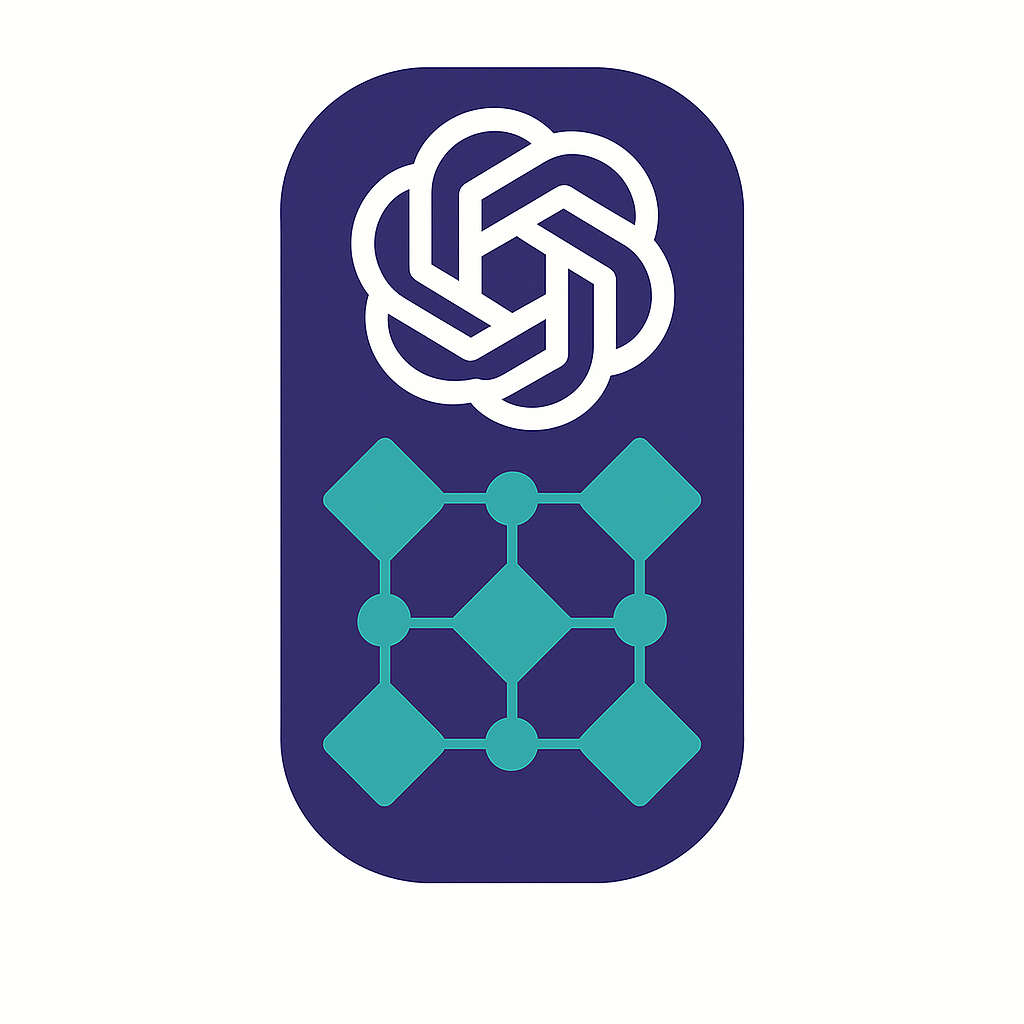}}

\usepackage{microtype}

\title{\perologo{}Perovskite-LLM: Knowledge-Enhanced Large Language Models for Perovskite Solar Cell Research}

\author{
  Xiang Liu$^{1,*}$ \quad 
  Penglei Sun$^{1,*}$ \quad 
  Shuyan Chen$^{1,2}$ \quad 
  Longhan Zhang$^{1,2}$ \\
  \textbf{Peijie Dong}$^1$ \quad 
  \textbf{Huajie You}$^1$ \quad 
  \textbf{Yongqi Zhang}$^1$ \quad 
  \textbf{Chang Yan}$^{1,2}$ \\
  \textbf{Xiaowen Chu}$^{1,2,\dagger}$ \quad 
  \textbf{Tong-yi Zhang}$^{1,2,\dagger}$ \\
  $^1$The Hong Kong University of Science and Technology (Guangzhou) \\
  $^2$Guangzhou Municipal Key Laboratory of Materials Informatics \\
}

\begin{document}
\maketitle
\begin{abstract}
The rapid advancement of perovskite solar cells (PSCs) has led to an exponential growth in research publications, creating an urgent need for efficient knowledge management and reasoning systems in this domain. We present a comprehensive knowledge-enhanced system for PSCs that integrates three key components. First, we develop \textbf{Perovskite-KG}, a domain-specific knowledge graph constructed from 1,517 research papers, containing 23,789 entities and 22,272 relationships. Second, we create two complementary datasets: Perovskite-Chat, comprising 55,101 high-quality question-answer pairs generated through a novel multi-agent framework, and Perovskite-Reasoning, containing 2,217 carefully curated materials science problems. Third, we introduce two specialized large language models: \textbf{Perovskite-Chat-LLM} is used for domain-specific knowledge assistance, and \textbf{Perovskite-Reasoning-LLM} is used for scientific reasoning tasks. Experimental results demonstrate that our system significantly outperforms existing models in domain-specific knowledge retrieval and scientific reasoning tasks, providing researchers with effective tools for literature review, experimental design, and complex problem solving in PSC research.
\end{abstract}
\def\thefootnote{$*$}\footnotetext{Equal Contribution.}

\def\thefootnote{$\dagger$}\footnotetext{Corresponding Author.}

\section{Introduction}
Perovskite solar cells (PSCs) have emerged as one of the most promising next-generation photovoltaic technologies, achieving remarkable progress with power conversion efficiencies (PCEs) exceeding 27.0\% within just over a decade \cite{NREL2025,snaith2018present,correa2017promises,wu2021main,ang2022comprehensive,sathaye2011renewable,bogdanov2019radical}. The rapid development of PSCs has generated an exponential growth in research publications, making it increasingly challenging for researchers to efficiently access and utilize the vast amount of knowledge in this field. This challenge is particularly acute given the complex interplay between material composition, fabrication processes, and device structure that characterizes PSC research.

Traditional approaches to scientific knowledge management, such as literature reviews and databases, while valuable, are limited in their ability to capture the intricate relationships between different aspects of PSC research~\cite{yang2024achievements,han2025perovskite}. Furthermore, existing artificial intelligence systems in materials science typically focus on specific prediction tasks or general scientific knowledge, lacking the specialized capability to handle the unique characteristics of the research of perovskite solar cells and more general materials science research~\cite{han2025perovskite,seifrid2022autonomous,mishra2024foundational,song2023matsci, schilling2025text,miret2024llms}. This gap highlights the need for an integrated system that can both systematically organize domain knowledge and provide intelligent assistance to researchers.

To address these challenges, we present a comprehensive knowledge-enhanced system specifically designed for the perovskite solar cell domain, consisting of three key components. First, we develop \textbf{Perovskite-KG}, a domain-specific knowledge graph constructed from 1,517 research articles, containing 23,789 entities and 22,272 relationships between manufacturing processes, parameters, and performance metrics. Second, we create a multi-agent framework for generating high-quality instruction-tuning data, which not only reduces annotation costs but also ensures high reliability and low hallucination through the synergy of multiple specialized agents and expert guidance. This framework generates two complementary datasets: (1) \textbf{Perovskite-Chat}, an instruction-tuning dataset comprising 55,101 high-quality question-answer pairs generated from 2,214 high-impact papers using a novel multi-agent framework, and (2) \textbf{Perovskite-Reasoning}, a collection of 2,217 carefully curated materials science problems designed to enhance scientific reasoning capabilities. Third, we introduce two specialized large language models: \textbf{Perovskite-Chat-LLM} for domain-specific knowledge assistance and \textbf{Perovskite-Reasoning-LLM} for tackling complex materials science reasoning tasks. Our work makes the following key contributions:

\noindent
\textbf{\ding{182} Construction of a comprehensive perovskite knowledge graph.}
We construct the first comprehensive knowledge graph for perovskite solar cells, organizing domain knowledge into a structured format that captures the relationships between materials, processes, and device performance.

\vspace{2pt}
\noindent
\textbf{\ding{183} Multi-agent framework for high-quality data generation.}
We propose an effective multi-agent framework for generating high-quality instruction-tuning data, resulting in two specialized datasets: a diverse domain-specific dataset covering seven research categories and a focused reasoning dataset for enhancing scientific problem solving capabilities.

\vspace{2pt}
\noindent
\textbf{\ding{184} Specialized LLMs for perovskite research.}
We develop and evaluate two specialized large language models for perovskite solar cells that demonstrate superior performance compared to baseline models: one optimized for domain-specific queries and another for scientific reasoning tasks.

\vspace{2pt}
\noindent
\textbf{\ding{185} Extensive experimental validation.}
We provide extensive experimental results showing the effectiveness of our integrated system in supporting various research tasks, from literature review to experimental design and complex problem solving in materials science.

\section{Related Work}

\subsection{LLM in Materials Science}
The convergence of language modeling and computational materials science has unlocked transformative potential for rapid discovery. Recent breakthroughs in domain-specific architectures (e.g. hierarchical attention mechanisms \cite{kononova2021opportunities} and multimodal fusion networks \cite{swain2016chemdataextractor}) have addressed critical challenges in crystal structure prediction \cite{walker2021impact} and phase diagram analysis \cite{TREWARTHA2022100488}. As evidenced by the Materials Genome Initiative benchmarks \cite{Tshitoyan2019}, three main research thrusts have emerged: (1) structured information extraction from heterogeneous corpora, (2) knowledge graph embeddings for composition-property relationships, and (3) neurosymbolic reasoning for synthesis pathway optimization~\cite{seifrid2022autonomous,mishra2024foundational,song2023matsci, schilling2025text,miret2024llms}.

Building upon these foundations, knowledge-enhanced systems have achieved state-of-the-art performance through two complementary paradigms: graph-based approaches employing heterogeneous graph neural networks (HGNNs) now attain 89.7\% accuracy on multi-hop material property queries \cite{an2024knowledgegraphquestionanswering}, while agent-based frameworks demonstrate 18.7\% improvement in autonomous experimental design through chain-of-thought prompting \cite{zhang-etal-2024-honeycomb, song2023honeybee, gupta2022matscibert,hira2024reconstructing}.

The field's maturation is further evidenced by systematic resource development: (i) The SciQAG framework \cite{wan2024sciqagframeworkautogeneratedscience} introduces a novel curriculum learning paradigm for generating 120K domain-specific QA pairs, reducing expert annotation requirements by 78\%; (ii) Standardized evaluation now spans chemical synthesis (ChemLLMBench's reaction yield prediction task \cite{guo2023largelanguagemodelschemistry}), biomedical applications (MultiMedQA's toxicity prediction challenge \cite{MultiMedQA}), and cross-domain reasoning (SciEval's materials-device co-design track \cite{sun2023scieval}).

\subsection{Knowledge Graph in Materials Science}
Domain-specific knowledge graphs have evolved into structured semantic frameworks that systematically consolidate heterogeneous multi-source data through machine-readable representations, allowing cross-domain knowledge integration to accelerate discovery pipelines~\cite{pan2024unifying,song2024scene,zhu2022multi,bordes2013translating,galkintowards}. 
In materials informatics, current implementations manifest two distinct paradigms: literature-derived systems exemplified by MatKG~\cite{venugopal2024matkg} and DISCOMAT~\cite{gupta2023discomat}, which employ NLP and graph techniques to extract material compositions from textual sources, while empirical architectures represented by MatSciKB~\cite{zhang2024honeycomb}, Propnet~\cite{mrdjenovich2020propnet}, MekG~\cite{statt2023materials}, and MOF-KG~\cite{an2022building} focus on encoding experimental provenance and computational models through graph-based representations of material lineages. 
However, these approaches face the challenges that manual curation processes face with resource burdens, while existing extraction methods exhibit limited granularity in resolving complex synthesis-process-property relationships from unstructured text. 
To address these limitations, we propose an LLM-driven framework specifically optimized for perovskite materials research that features a hybrid architecture that synergizes domain ontologies with self-supervised relationship extraction, augmented by automated quality control pipelines that enforce materials science constraints. 

\subsection{Multi-agent systems}
The landscape of AI system architectures encompasses two distinct paradigms: multi-agent systems and autonomous agents ~\citep{zhuge2023mindstorms, hong2024data, jia2024mobile, wang2023voyager,su2025cafes}. While autonomous agents rely on independent decision-making capabilities, multi-agent systems excel through structured collaboration between specialized components. The latter approach offers practical advantages by building upon established expertise rather than requiring complex behavioral modeling.

Research in multi-agent frameworks has evolved along two primary trajectories. The first focuses on domain-agnostic systems that leverage collective intelligence for general problem-solving ~\citep{wei2022COT, diao2023active, wang2022COTSC, madaan2023self, wang2023multipersona}. The second pathway explores domain-specific applications, with notable implementations in: Code generation and debugging ~\citep{sirui2024meta, Tal2024Alpha, li2024debug}, Data analytics ~\citep{yu2024hai, yi2024gen, li2024dawn, zhou2023llm}, Mathematical reasoning ~\citep{zhong2024achieving, xu2024lemur,pan-etal-2024-plum}, Knowledge retrieval ~\citep{nori2023medprompt, zhou2024language}

Despite significant progress in identifying effective agent configurations for specific use cases, the field still faces the challenge of developing systematic approaches for new domains. This highlights the importance of research into automated methods for framework design and optimization.

In this section, we collect $1,517$ paper in perovskite domain to build Perovskite-KG and design the automatic knowledge graph construction pipeline including three stages document filtering, knowledge extracting and knowledge graph organization~\cite{mrdjenovich2020propnet}, as shown in the Appendix ~\ref{appendix:schema_in_Perovskite_KG}.

\subsection{Reasoning alignment}
Recent advances in parameter-efficient alignment have witnessed multiple research teams pursue different methodologies to align the performance of o1~\citep{o1}. Contemporary approaches bifurcate along two technical axes: (1) reinforcement learning paradigms exemplified by DeepSeek-R1's adversarial preference optimization \citep{guo2025deepseek} and K1.5's multi-objective reward shaping \citep{k1.5}, versus (2) supervised fine-tuning strategies employing distilled datasets at scale ($\geq10^4$ examples) as demonstrated in \citep{sky_t1,xu2025redstardoesscalinglongcot,bespoke_stratos}. In particular, S1~\citep{s1} and LIMO~\citep{limo} operationalize the Superficial Alignment Hypothesis \citep{zhou2023lima} through curriculum-based sparse fine-tuning, achieving comparable reasoning capabilities with merely 1,000-2,000 carefully curated examples, a 92\% reduction in annotation costs relative to conventional SFT approaches.

\section{Perovskite-KG}

\begin{figure*}[t]
  \centering
  \includegraphics[scale=0.4,trim=0 0 0 0,clip]{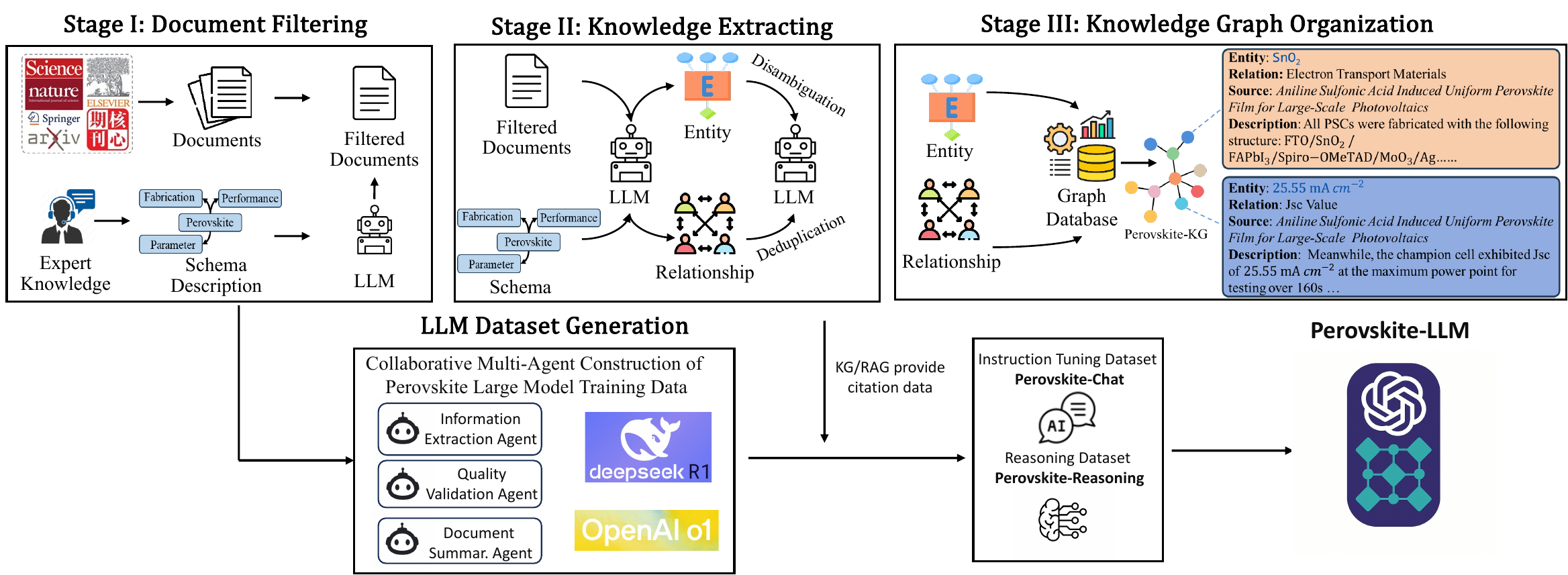}
    \caption{The pipeline of Perovskite-KG construction and Perovskite-LLM.}
    \vspace{-15pt}
  \label{fig:kg_pipeline}
\end{figure*}

\noindent $\bullet$ \textbf{Document Filtering}. 
Drawing upon expert knowledge, we have developed the schema for perovskite materials. 
This schema, shown in the Appendix Table~\ref{tab:schema_kg}, integrates three ontologies $\{o_i \mid o_i \in \text{schema} \}$: fabrication, parameters, and performance.
The fabrication ontology encompasses the procedures and conditions required to synthesize perovskite materials. 
The parameters ontology defines the ingredients, structural components, and other compositional aspects of the device. 
The performance ontology is concerned with the efficiency and functional characteristics of perovskite devices.
Each ontology $o_i$ is further divided into sub-ontologies $so_i^{(j)}$, where $o_i = \bigcup_{j=1}^{n_i} so_i^{(j)}$ and $n_i$ represents the number of sub-ontologies within $o_i$. 
Each sub-ontology $so_i^{(j)}$ provides a domain-specific description, denoted as $d_i^{(j)}$, along with a corresponding data type, denoted as $t_i^{(j)}$, that is relevant to its particular scope.

For each sub-ontology $[ so_i^{(j)}, d_i^{(j)}, t_i^{(j)} ]$ (e.g., \textit{"Coating Parameter" - "Details about the coating method used in the material deposition process" - "Float"}), we create the prompts to query documents $D = \{ D_k \mid k = 1, \dots, m \}$ using a large language model.
These prompts facilitate the extraction of relevant information for each sub-ontology. 
The output $D_{\text{filtered}}^{(i, j)}$ is defined as:

\begin{equation}
\begin{aligned} 
D_{\text{filtered}}^{(i, j)} = \{ D_k \in D \mid so_i^{(j)} \subset D_k \},
\end{aligned}
\end{equation}
where $D_{\text{filtered}}^{(i, j)}$ represents the set of filtered documents containing pertinent details for sub-ontology $so_i^{(j)}$ across the collection. 
This approach ensures a systematic and efficient retrieval of targeted information for each sub-ontology.

\noindent $\bullet$ \textbf{Knowledge Extracting}.
We employ a prompt function, denoted as $f_{\text{prompt}}(\cdot)$, to transform the sub-ontology $[so_i^{(j)}, d_i^{(j)}, t_i^{(j)}]$ into a document prompt, represented as $f_{\text{prompt}}(so_i^{(j)}, d_i^{(j)}, t_i^{(j)})$. 
To extract the potential domain knowledge $K$, we utilize a pre-trained large language model (LLM), expressed as $\text{LLM}(\cdot; \theta)$, under a zero-shot setting where the parameters $\theta$ remain fixed.
The whole pipeline can be formulated as below:
\begin{equation}
\begin{aligned}
\scalebox{0.92}{$
K = \underset{D_{\text{filtered}}^{(i, j)}}{\text{search}} \; \; \text{LLM}(f_{\text{prompt}}(so_i^{(j)}, d_i^{(j)}, t_i^{(j)}); \theta),
$}
\end{aligned}
\end{equation}
where the search function $\text{search}(\cdot)$ may involve an argmax operation to identify the highest-scoring output or a sampling approach to generate outputs according to the probability distribution specified by the adopted $\text{LLM}(\cdot; \theta)$. 

After extracting knowledge, we conduct quality control procedures to ensure accuracy and reliability. These procedures include entity disambiguation and relationship deduplication.
Entity disambiguation in a knowledge graph aims to resolve ambiguity by identifying the unique entity that corresponds to an ambiguous mention, denoted as $e_{\text{mention}}$, within a subgraph. 
The objective is to determine a distinct entity $e^*$ that accurately represents $e_{\text{mention}}$.
Relationship deduplication involves identifying and merging redundant relations in the knowledge graph. For instance, given two relations $r_i = (e_1, r, e_2)$ and $r_j = (e_1', r', e_2')$, if they convey the same semantic meaning—that is, if $(e_1, e_2)$ and $(e_1', e_2')$ refer to identical entities and the relations $r$ and $r'$ are equivalent.

\noindent $\bullet$ \textbf{Knowledge Graph  Organization}.
We construct the Perovskite Knowledge Graph (Perovskite-KG) using a graph database. The Perovskite-KG consists of $23,789$ entities and $22,272$ relationships. By incorporating citation relationships between papers, we enable our LLM to provide references for its responses, enhancing credibility and reducing hallucination.

\section{Instruction Tuning Dataset Generation}

In this section, we collect $2,214$ the top level publications papers in the perovskite domain and design the instruction tuning dataset including question answering and multiple choice questions, containing $55,101$ instances around 4.4 million tokens, named \textbf{Perovskite-Chat}. Our experiments show that our perovskite instruction tuning dataset can effectively improve the performance of LLMs on perovskite related tasks.

Figure \ref{fig:kg_pipeline} illustrates this multi-agent framework for the generation of  instruction tuning dataset. The process begins with expert guidance and academic literature from various sources (including Science, Nature, Elsevier, Springer, arXiv, and others) as input. Expert guidance is provided by the domain expert focused on 7 research categories, 21 research questions. Table \ref{tab:question_classification} further expands this classification by presenting 21 specific research questions (Q1-Q21) organized within these seven categories; more details can be found in Appendix \ref{appendix:dataset_statistics}. These inputs feed into a multi-agent system: (1) an Information Extraction Agent that processes the raw content, (2) a Quality Validation Agent that ensures data accuracy and relevance, and (3) a Document Summarization Agent that condenses and structures the information.  This framework ensures systematic, high-quality data processing through multiple validation and refinement stages. 

Let $D = \{d_1, ..., d_n\}$ represent the collection of academic literature from various sources, and $E = \{c_1, ..., c_7\}$ denote the expert guidance categories with the corresponding research questions $Q = \{q_1, ..., q_{21}\}$. The multi-agent framework processes these inputs through three specialized agents:

Information Extraction: 
\begin{equation}
A_{\text{extract}}(d_i) = \{x_1, ..., x_k\}
\end{equation}

Quality Validation: 
\begin{equation}
A_{\text{validate}}(x_j) = \begin{cases} 
1, & \text{if valid} \\
0, & \text{otherwise}
\end{cases}
\end{equation}

Document Summarization: 
\begin{equation}
A_{\text{summarize}}(X_{\text{valid}}) = y
\end{equation}

The final instruction tuning dataset $\mathcal{D}$ is constructed as:

\begin{equation}
\mathcal{D} = \{(q_i, y_i) \mid q_i \in Q, \nonumber
\end{equation}
\begin{equation}
y_i = A_{\text{summarize}}(A_{\text{validate}}(A_{\text{extract}}(d_i)))\}
\end{equation}

\begin{table*}[t]
  \centering
  \begin{tabular}{p{3cm}|p{11cm}}
  \toprule
  \textbf{Category} & \textbf{Rationale} \\
  \midrule
  Device Structure  & Fundamental aspects focusing on high-efficiency (\textgreater 25\% PCE) device architecture and fabrication processes (Q1-Q3) \\
  \midrule
  Perf. Enhancement  & Analysis of problem-solving approaches and strategic choices in high-performance devices (Q4-Q5) \\
  \midrule
  Metrics  & Key performance indicators (V\textsubscript{OC}, FF, J\textsubscript{SC}) and their optimization methods (Q6-Q9) \\
  \midrule
  Stability  & Critical stability aspects addressing main degradation pathways: moisture, thermal, and light stability (Q10-Q12) \\
  \midrule
  Defect \& Recom.  & Fundamental mechanisms affecting device efficiency through defect passivation and recombination control (Q13-Q14) \\
  \midrule
  Interface  & Interface engineering and charge transport optimization (Q15-Q17) \\
  \midrule
  Materials  & Comprehensive analysis of functional materials and their characteristics in different device components (Q18-Q21) \\
  \midrule
  \end{tabular}
    \vspace{-10pt}

  \caption{Classification of Research Questions in Perovskite Solar Cell Studies}
  \vspace{-0.4cm}
  \label{tab:question_classification}
\end{table*}

\begin{figure}[!t]
  \centering
  \subfigure[The distribution of question categories in the instruction tuning dataset.]{
      \includegraphics[width=0.9\linewidth]{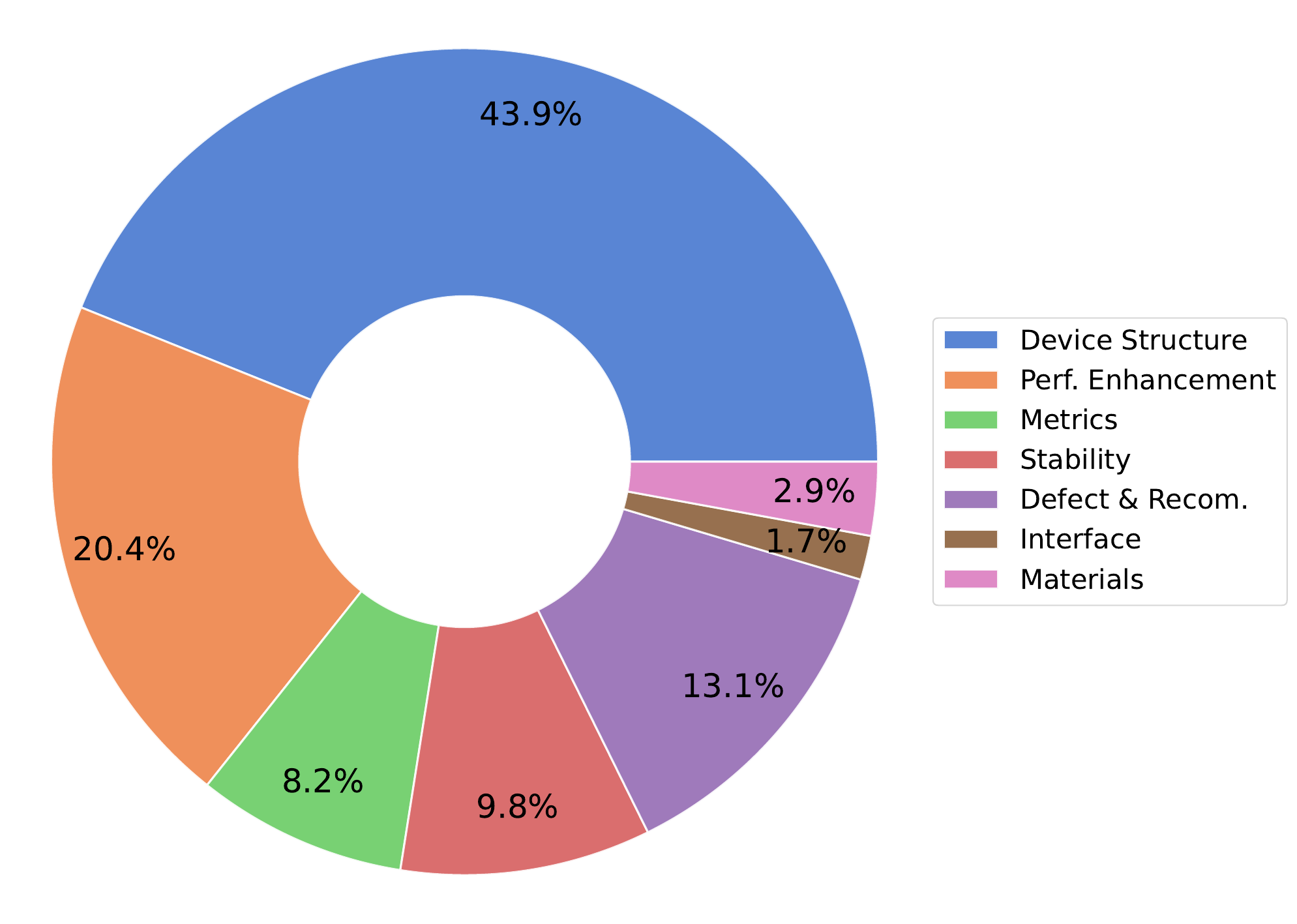}
  }
  \subfigure[The word cloud of the instruction tuning dataset.]{
      \includegraphics[width=1\linewidth]{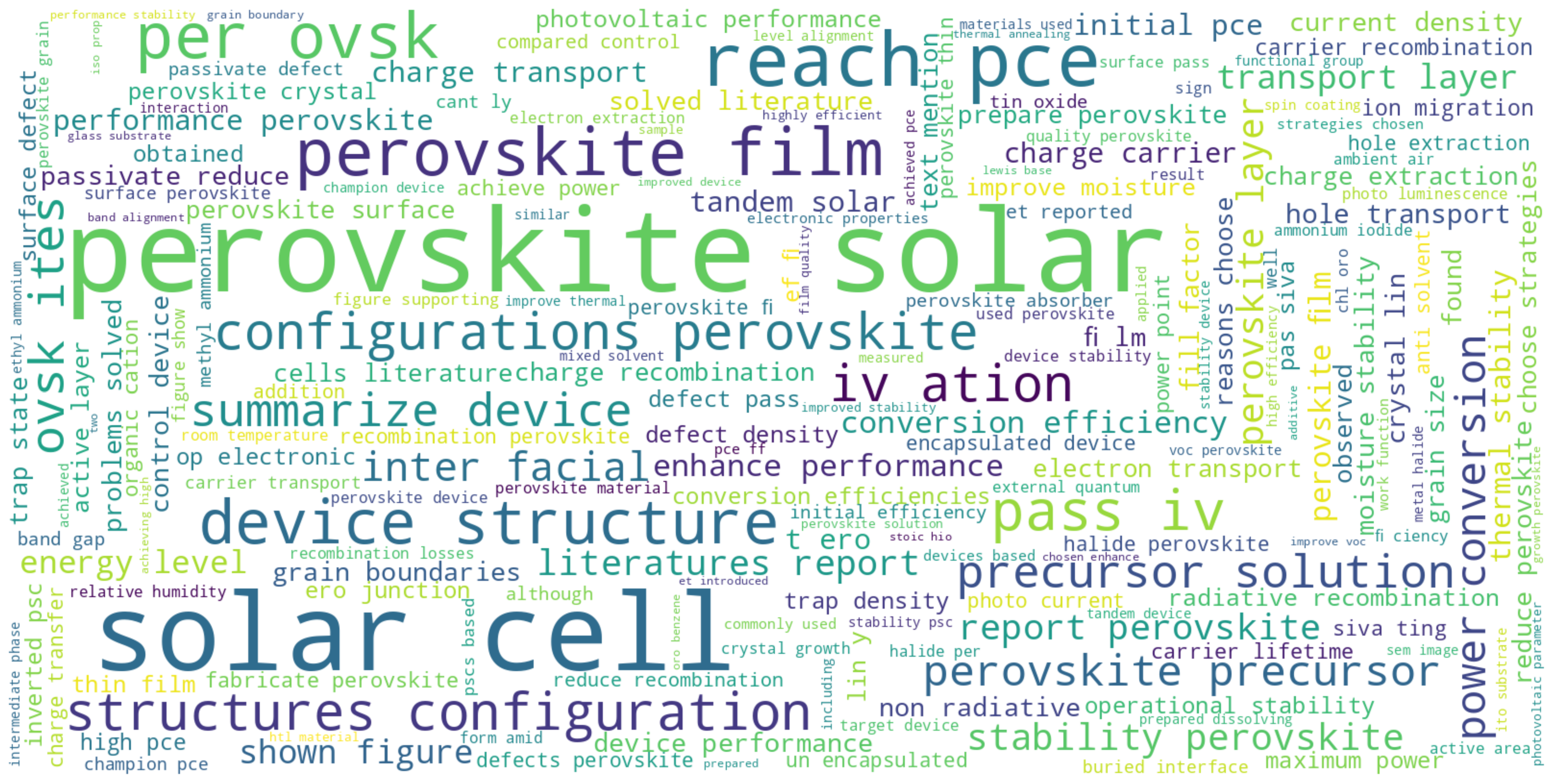}
  }
    \vspace{-10pt}

  \caption{The distribution of question categories in the instruction tuning dataset.}
  \vspace{-15pt}
  \label{fig:question_categories_distribution}
\end{figure}

Next, we introduce \textbf{Perovskite-Reasoning}, a collection of 2,217 high-quality questions from materials science textbooks, designed to enhance reasoning capabilities in the perovskite and materials science domains. The questions were sourced from hundreds of widely used materials science and engineering textbooks, with a focus on perovskite solar cells and fundamental materials science concepts. Our rigorous selection process applied three key criteria: clarity of problem statements, completeness of solution, and alignment with core materials science principles. Materials science professors conducted expert assessments to categorize questions by difficulty level, validated through student performance data and baseline model testing. To develop comprehensive solution paths, we employed advanced language models like DeepSeek-R1~\cite{guo2025deepseek} and O1~\cite{o1} in a multi-step reasoning approach. This methodology involved decomposing complex problems into logical steps, applying key physical and chemical principles, and implementing systematic solution strategies with result validation. The resulting dataset features detailed reasoning chains that demonstrate step-by-step problem-solving processes, making it valuable for training models in scientific reasoning and materials science problem-solving.

\paragraph{Training Dataset}

Figure \ref{fig:question_categories_distribution} (a) presents the distribution of the question categories in the \textbf{Perovskite-Chat} instruction tuning data set. Device Structure dominates with 43.9\% of all questions, followed by Performance Enhancement at 20.4\%. Device \& Recom. comprises 13.1\%, while Metrics accounts for 8.2\%. The remaining categories include Stability (9.8\%), Materials (2.9\%), and Interface (1.7\%). Figure \ref{fig:question_categories_distribution} (b) displays a wordcloud visualization of the dataset's most frequent terms, with "perovskite solar" and "solar cell" appearing as predominant phrases, reflecting the dataset's core focus. Other frequently occurring terms such as "device structure," "configuration," and "stability" underscore the key technical aspects addressed. This distribution demonstrates the dataset's strong emphasis on device structural aspects, with performance-related queries forming the second largest category.

\paragraph{Evaluation Dataset}
For better evaluation, we design the evaluation dataset including multiple choice questions and question answering in the perovskite domain. The evaluation dataset also extract from the top level publications in perovskite domain with our multi-agent framework and extral expert double check. The evaluation dataset contains 1,103 question answering named \textbf{Perovskite QA} and 1,103 multiple choice questions named \textbf{Perovskite MCQ}.

For question answering, we set the Rouge-L score and the LLM-as-a-Judge~\citep{zheng2023judging} score as the evaluation metric. In our experiments, we find that both metrics can effectively measure the quality of question answering and consistency with each other.

For multiple choice questions, we set the accuracy as the evaluation metric. Using LLaMA-3.1-8B-Instruct~\citep{llama3} as the baseline model, the difficulty level of each question is determined by its zero shot performance. Specifically, if LLaMA-3.1-8B-Instruct can correctly answer a question in a zero-shot setting (without any task-specific training or prompt engineering), we classify it as an "easy" question. Conversely, questions that LLaMA-3.1-8B-Instruct fails to answer correctly are categorized as "Hard". This classification method resulted in 823 easy questions and 280 hard questions in our evaluation dataset, providing a balanced assessment of model capabilities across different difficulty levels.

To evaluate the performance of \textbf{Perovskite-Reasoning}, we incorporated Minerva~\cite{lewkowycz2022solving} and GPQA Diamond~\cite{rein2023gpqa} as a benchmark. These contain undergraduate and PhD level science questions from biology, chemistry, and physics. 
\section{Perovskite-LLM}
\subsection{Experiment Design}
In this section, we conduct the instruction tuning experiments on the \textbf{Perovskite-Chat} and \textbf{Perovskite-Reasoning} dataset. We select the LLaMA-3.1-8B-Instruct~\citep{llama3} and Qwen-2.5-7B-Instruct~\citep{yang2024qwen2} as the baseline model, and \textbf{Perovskite-Chat-LLM} and \textbf{Perovskite-Reasoning-LLM} are fine-tuned version of LLaMA-3.1-8B-Instruct and Qwen-2.5-7B-Instruct with \textbf{Perovskite-Chat} and \textbf{Perovskite-Reasoning} dataset. For the training methodology, we employed a full fine-tuning approach. While Parameter-Efficient Fine-Tuning (PEFT) methods like LoRA~\citep{hu2021lora} are prominent for their efficiency, some studies suggest they can exhibit weaker convergence on downstream tasks compared to full fine-tuning~\citep{pan2024lisa}. As our computational resources were sufficient, we opted for full fine-tuning to maximize model performance and adaptation to our specialized datasets. We leave the exploration of PEFT methods for future work, especially when scaling to larger models.

For the training process, we used the full parameter fine-tuning method to fine-tune the Perovskite-LLM. The experiment is carried out on the A800 GPU server, with flash attention~\citep{dao2023flashattention2} and mixed precision training for efficient training. For more details on the training process, please refer to Appendix \ref{appendix:instruction_tuning}.

For the evaluation process, we use the perplexity (PPL), Rouge-L score, LLM-Judge, and Human-Judge score to evaluate the performance on the Perovskite QA benchmark, the accuracy to evaluate the performance on the Perovskite MCQ benchmark, and the pass@1 rate on Minerva and GPQA benchmarks to evaluate the performance on the Perovskite-Reasoning-LLM. All experiments are conducted with zero-shot settings and three times to get the average results.

\subsection{Results and analysis}
\paragraph{Perovskite-Chat-LLM}
Table \ref{tab:val_qa_scores} presents the evaluation results of various models on the Perovskite QA dataset. The baseline models include GPT-3.5-Turbo, GPT-4o-Mini, GPT-4o, LLaMA-3.1-8B. Among these, Perovskite-Chat-LLM demonstrates superior performance across all metrics, achieving a perplexity of 2.97, a Rouge-L score of 41.25, and an LLM-Judge score of 2.97. This represents a significant improvement over the baseline LLaMA-3.1-8B model. The GPT family of models, while competitive in terms of LLM-Judge scores, showed lower performance in Rouge-L scores compared to Perovskite-Chat-LLM, with GPT-4o achieving 11.36 for Rouge-L and 1.41 for LLM-Judge. With the case study in Figure \ref{fig:example_perovskite_llm_chatgpt}, we can see that Perovskite-Chat-LLM can generate more accurate and consistent answers compared to other models, and ChatGPT only can offer a general and nonspecific answer which leads to low performance on the Rouge-L and LLM-Judge metrics.

To further validate the effectiveness and credibility of our model, we incorporated human expert evaluation into our assessment methodology. Table~\ref{tab:human_expert_eval} presents the results of this human expert evaluation alongside the automatic metrics. Notably, Perovskite-Chat-LLM achieved the highest scores in both LLM-Judge (2.97) and Human Expert (3.12) assessments, outperforming baseline models such as GPT-4o and LLaMA-3.1-8B. This consistency across both human and LLM-Judge evaluations demonstrates the robustness and reliability of our model, as well as its ability to generate responses that are not only accurate but also align with expert expectations in the field.

\begin{table}[t]
  \centering
  \resizebox{\columnwidth}{!}{
  \begin{tabular}{l|cc}
    \toprule
    \textbf{Model} &  \textbf{LLM-Judge} $\uparrow$ & \textbf{Human Expert} $\uparrow$ \\
    \midrule
    GPT-4o &  1.41 & 1.73 \\
    LLaMA-3.1-8B & 1.28 & 1.17 \\
    \rowcolor{red!20} \textbf{Perovskite-Chat-LLM}  & \textbf{2.97} & \textbf{3.12} \\
    \bottomrule
  \end{tabular}
  }
    \vspace{-10pt}

  \caption{Comparison of automatic and human expert evaluation scores on the Perovskite QA dataset.}
  \label{tab:human_expert_eval}
\end{table}

\begin{table}[t]
  \centering
  \resizebox{\columnwidth}{!}{
  \begin{tabular}{l|ccc}
    \toprule
    \multirow{2}{*}{\textbf{Model}} & \multicolumn{3}{c}{\textbf{\textit{Perovskite QA}}} \\
    & \textbf{PPL} $\downarrow$ & \textbf{Rouge-L} $\uparrow$ & \textbf{LLM-Judge} $\uparrow$ \\
    \midrule
    GPT-3.5-Turbo & - & 11.24 & 1.24 \\
    GPT-4o-Mini & - & 11.90 & 1.34 \\
    GPT-4o & - & 11.36 & 1.41 \\
    \midrule
    LLaMA-3.1-8B & 6.77 & 13.18 & 1.28 \\
    LLaMA-3.1-70B & 4.98 & 17.38 & 1.80 \\
    Qwen-2.5-7B & 6.23 & 11.22 & 1.39 \\
    Qwen-2.5-72B & 5.12 & 10.17 & 1.31 \\
    \rowcolor{red!20} \textbf{Perovskite-Chat-LLM} & \textbf{2.97} & \textbf{41.25} & \textbf{2.97} \\
    \rowcolor{cyan!10} \textbf{       +Perovskite-KG} & \textbf{2.80} & \textbf{43.10} & \textbf{3.15} \\
    \bottomrule
  \end{tabular}
  }
  \vspace{-5pt}

  \caption{Performance of \textbf{Perovskite-Chat-LLM} on \textit{Perovskite QA}}
  \label{tab:val_qa_scores}
\end{table}

Table \ref{tab:val_mcq_scores} presents the evaluation results of various models on the Perovskite MCQ dataset, categorized into Easy, Hard, and All difficulty levels. Among these, GPT 4o achieves the highest overall score of 84.68, with scores of 91.37 for Easy and 65.00 for Hard questions. Perovskite-Chat-LLM, highlighted in red, shows strong performance with a score of 62.86 on Hard questions, the highest in this category, and an overall score of 87.22. 

\begin{table}[h]
  \centering

  \resizebox{0.9\columnwidth}{!}{
  \begin{tabular}{l|ccc}
    \toprule
    \multirow{2}{*}{\textbf{Model}} & \multicolumn{3}{c}{\textbf{\textit{Perovskite\_MCQ}}} \\
    & \textbf{Easy} & \textbf{Hard} & \textbf{All} $\uparrow$ \\
    \midrule
    GPT-3.5-Turbo & 86.63 & 49.29 & 77.15 \\
    GPT-4o-Mini & 89.79 & 61.79 & 82.68 \\
    GPT-4o & 91.37 & 65.00 & 84.68 \\
    \midrule
    \rowcolor{gray!20} LLaMA-3.1-8B & 100.00 & 0.00 & 74.21 \\
    LLaMA-3.1-70B & 93.44 & 66.43 & 86.58 \\
    Qwen-2.5-7B & 92.22 & 55.36 & 82.86 \\
    Qwen-2.5-72B & 93.07 & 64.29 & 85.77 \\
    \rowcolor{red!20} \textbf{Perovskite-Chat-LLM} & \textbf{95.50} & \textbf{62.86} & \textbf{87.22} \\
    \rowcolor{cyan!10} \textbf{       +Perovskite-KG} & \textbf{97.32} & \textbf{65.10} & \textbf{89.21} \\    \bottomrule
  \end{tabular}
  }
    \vspace{-5pt}

  \caption{Performance of \textbf{Perovskite-Chat-LLM} on \textit{Perovskite MCQ}. The LLaMA-3.1-8B baseline model's performance defines Easy/Hard question categories.}
  \vspace{-0.4cm}
  \label{tab:val_mcq_scores}
\end{table}

\paragraph{Perovskite-Reasoning-LLM}
Table \ref{tab:reasoning_benchmark} presents the evaluation results of Perovskite-Reasoning-LLM compared against various baseline models on the GPQA and Minerva benchmarks. In the 7B-scale model category, our Perovskite-Reasoning-LLM achieves state-of-the-art performance with remarkable data efficiency over 7B models, requiring only 2K training examples compared to 800K for R1-Qwen2.5-7B and 114K for OpenThinker-7B. Our model achieves 43.95 on GPQA and 44.49 on Minerva. Compared to 32B models, while our GPQA performance shows room for improvement (suggesting GPQA's sensitivity to model size), our Minerva score (44.49) is competitive with larger models like LIMO-32B (44.90) and approaches S1-32B (47.79). This shows that our efficient training approach can achieve strong performance on STEM reasoning tasks even with a smaller model architecture.

\begin{table}[htbp]
  \centering
  \resizebox{\columnwidth}{!}{
  \begin{tabular}{l|ccc|c}
    \toprule
    \textbf{Model} & $\#$ ex & \textbf{GPQA} $\uparrow$ & \textbf{Minerva} $\uparrow$ & \textbf{Avg} $\uparrow$ \\
    \midrule
    \multicolumn{4}{c}{API Models} \\
    \midrule
    o1 & - & 77.30 & - & - \\
    o1-preview & - & 73.30 & 47.10 & 60.20 \\
    o1-mini & - & 60.00 & - & - \\
    Deepseek-R1 & - & 71.50 &  - & - \\ 
    \midrule
    \multicolumn{4}{c}{32B} \\
    \midrule
    Qwen2.5-32B-Instruct & - & 48.00 & 41.20 & 44.60 \\
    QwQ-32B-preview & - & 65.10 & 39.00 & 52.05 \\
    LIMO-32B* & 0.8K & 66.70 & 44.90 & 55.80 \\
    S1-32B* & 1K & 59.60 & 47.79 & 53.69 \\
    \midrule
    \multicolumn{4}{c}{7B} \\
    \midrule
    LLaMA-3.1-8B Instruct & - & 30.40 & 13.97 & 22.18 \\
    R1-Qwen2.5-7B* & 800K & \textbf{44.49} & 25.25 & 34.87 \\
    R1-LLaMA3-8B* & 800K & 19.19 & 30.51 & 24.85 \\
    OpenThinker-7B* & 114K & 42.90 & \underline{41.10} & 42.00 \\
    \rowcolor{red!20} \textbf{Perovskite-R-LLM} & 2K& \underline{43.95} & \textbf{44.49} & \textbf{44.22} \\ 
    \bottomrule
  \end{tabular}
  }
    \vspace{-10pt}

  \caption{We evaluate the performance of Perovskite-Reasoning-LLM on the GPQA and Minerva benchmarks.
  * indicates the results are from our evaluation. $\#$ ex = number of examples used for fine-tuning. $\#$ Perovskite-R-LLM : Perovskite-Reasoning-LLM}
  \vspace{-15pt}
  \label{tab:reasoning_benchmark}
\end{table}

\subsection{Integrate Perovskite-KG with Perovskite-Chat-LLM}

To ensure the Perovskit-Chat-LLM possesses current and specialized knowledge, we employ retrieval-augmented generation (RAG) to dynamically integrate information from our Perovskite-KG. 
A key advantage of this approach is the continuous and automatic updating of the LLM's knowledge base, which is directly facilitated by the automated construction methodology of the Perovskite-KG itself. 
The technical implementation of this integration leverages the LightRAG framework \cite{guo2024lightrag}.
The performance improvements resulting from the integration of Perovskite-KG with Perovskite-Chat-LLM ("\textit{+Perovskite-KG}") are presented in Table ~\ref{tab:val_qa_scores} and Table ~\ref{tab:val_mcq_scores}. 
When enhanced by the Perovskite-KG, the Perovskite-Chat-LLM demonstrates improved scores in both Perovskite QA ($2.97 \to 3.15$) and Perovskite MCQ ($87.22 \to 89.21$). 
Representative case studies illustrating these benefits are detailed in Figure~\ref{fig:perovskite_llm_cite_small} and Figure~\ref{fig:example_perovskite_llm_cite}.
The Perovskit-Chat-LLM, when augmented with this knowledge integration, demonstrates an enhanced capability to access up-to-date information and deliver fine-grained, domain-specific responses compared to its baseline version lacking such integration. For instance, when presented with a query such as, ``\textit{Can you tell me what is CuO? And give me 5 usage in perovskite?}'', the enhanced model provides the requested details, with each stated usage accompanied by a citation from the underlying knowledge source.

\vspace{-7pt}
\subsection{Case Study}
To illustrate the practical advantages of Perovskite-LLM over existing models, Figure \ref{fig:example_perovskite_llm_chatgpt} presents a comparative analysis of the responses of Perovskite-Chat-LLM and ChatGPT to a question about the fabrication of high-efficiency perovskite solar cells. The responses demonstrate a clear distinction in the depth and specificity of knowledge provided by each model. For instance, Perovskite-Chat-LLM specifies precise conditions for the HTL preparation (150°C for 15 minutes) and details a two-step spin-coating procedure (1,000 rpm for 10 seconds, followed by 6,000 rpm for 30 seconds) with specific process modifications like anisole addition. This level of detail reflects Perovskite-Chat-LLM's specialized training on domain-specific literature and its ability to provide practical, implementation-ready guidance for researchers in the field. 
For more details, please refer to Appendix \ref{appendix:more_case_study}.

\begin{figure}[!t]
  \centering
  \tikzstyle{every node}=[font=\small,scale=0.9]
  \begin{tikzpicture}
      \footnotesize
      \definecolor{chatcolor1}{HTML}{FFFACD} 
      \definecolor{shadecolor}{gray}{0.95}
      \fontfamily{cmss}\selectfont
  
      \node[align=left, text width=0.4\textwidth, fill=shadecolor, rounded corners=1mm, draw=black, line width=0.5mm, fill opacity=0.7, text opacity=1, anchor=north west] (node1) at (-2cm,0) {
          \textbf{Question:}\\
          How to fabricate the perovskite solar cells those can reach PCE over 25\%?
      };
      \node[align=left, text width=0.4\textwidth, fill=chatcolor1, rounded corners=1mm, draw=black, line width=0.5mm, fill opacity=0.7, text opacity=1, anchor=north east] (node2) at ($(node1.south -| {0.6\textwidth,0}) + (-4.5cm,-0.2cm)$) {
          \textbf{Perovskite-Chat-LLM:} \\
          ... \\
          For the hole transport layer (HTL), materials like PEDOT:PSS are commonly used. These materials are spin-coated onto the substrate and annealed at a temperature of \textbf{150°C} for \textbf{15 minutes}. The perovskite precursor solution is then prepared using a mixed cation and halide composition, and spin-coated onto the HTL. The spin-coating process involves a two-step procedure: first at \textbf{1,000 rpm} for \textbf{10 seconds}, followed by \textbf{6,000 rpm} for \textbf{30 seconds}, with anisole added during the second step to facilitate crystallization.\\
          ... \\
          };      
      
        \node[align=left, text width=0.4\textwidth, fill=chatcolor1, rounded corners=1mm, draw=black, line width=0.5mm, fill opacity=0.7, text opacity=1, anchor=north east] (node2) at ($(node2.south -| {0.6\textwidth,0}) + (-4.5cm,-0.2cm)$) {
          \textbf{ChatGPT:} \\
          ... \\
          Perovskite Material Quality: The quality of the perovskite material is crucial for achieving high efficiency. Use high-quality perovskite precursors and optimize the fabrication process to ensure uniform and defect-free perovskite films. \\
          ... \\
          };      
        
  \end{tikzpicture}
  \caption{Comparison of responses between Perovskite-LLM and ChatGPT: Perovskite-LLM provides detailed operational steps with specific parameters, while ChatGPT only offers general conceptual guidance.}
  \vspace{-0.4cm}
  \label{fig:example_perovskite_llm_chatgpt}
  \end{figure}

  \begin{figure}[!t]
    \centering
    \tikzstyle{every node}=[font=\small,scale=0.9]
    \begin{tikzpicture}
        \footnotesize
        \definecolor{chatcolor1}{HTML}{FFFACD} 
        \definecolor{shadecolor}{gray}{0.95}
        \fontfamily{cmss}\selectfont
    
        \node[align=left, text width=0.4\textwidth, fill=shadecolor, rounded corners=1mm, draw=black, line width=0.5mm, fill opacity=0.7, text opacity=1, anchor=north west] (node1) at (-2cm,0) {
            \textbf{Question:}\\
            Can you tell me what is CuO? And give me 5 usage in perovskite.
        };
        \node[align=left, text width=0.4\textwidth, fill=chatcolor1, rounded corners=1mm, draw=black, line width=0.5mm, fill opacity=0.7, text opacity=1, anchor=north east] (node2) at ($(node1.south -| {0.6\textwidth,0}) + (-4.5cm,-0.2cm)$) {
            \textbf{Perovskite-Chat-LLM:} \\
            ...
            \\
            1. \textbf{Hole Transport Material}: CuO acts as an effective hole transport material (HTM) due to its suitable work function and high hole mobility. It is often used in conjunction with other materials, such as $\bm{NiO_x}$, to enhance the performance of PSCs by facilitating efficient charge extraction [1].
            \\
            ...
            \\
            \textbf{References:} \\

            [1] Inorganic Materials as Hole Selective Contacts and Intermediate Tunnel Junction Layer for Monolithic Perovskite‐CIGSe Tandem Solar Cells, Wang, Yajie; Wenisch, Robert; Schlatman \\ 
            ...

            };      
        
    \end{tikzpicture}
    \caption{A case study of Perovskite-Chat-LLM's ability to provide detailed and accurate information with references.}
    \vspace{-0.8cm}
    \label{fig:perovskite_llm_cite_small}
    \end{figure}

\vspace{-7pt}
\section{Conclusion}
In this work, we present a comprehensive knowledge-enhanced system for perovskite solar cell research, integrating three key components: (1) Perovskite-KG, a domain-specific knowledge graph containing 23,789 entities and 22,272 relationships; (2) multi-agent system for generating two complementary datasets, Perovskite-Chat and Perovskite-Reasoning, designed for domain-specific knowledge assistance and scientific reasoning respectively; and (3) two specialized large language models that demonstrate superior performance in both knowledge retrieval and reasoning tasks. Our experimental results show significant improvements over existing models, with Perovskite-Chat-LLM achieving state-of-the-art performance on domain-specific tasks and Perovskite-Reasoning-LLM showing competitive performance on scientific reasoning benchmarks despite using substantially fewer training examples. The system provides researchers with effective tools to review the literature, design experiments, and solve complex problems in PSC research. 
Our work demonstrates the potential of LLMs to accelerate innovation and discovery in materials science by enabling more efficient knowledge access and reasoning capabilities.

\section{Limitations}
Despite the promising results, our current system has several limitations that warrant future investigation:

\begin{itemize}
    \item \textbf{Knowledge Coverage}: While our knowledge graph covers a substantial portion of the PSC literature, it may not capture all emerging research directions and novel experimental techniques. Future work should focus on developing mechanisms for automatic knowledge base expansion and updates to maintain its relevance.
    

    \item \textbf{Model Size Trade-offs}: While our 7B-parameter models achieve competitive performance, there might be certain complex tasks that benefit from larger model architectures, suggesting a potential trade-off between efficiency and capability. Future work will investigate model compression techniques and more efficient architectures.
\end{itemize}

To address these limitations, our future work will focus on three main directions: (1) developing a continuous knowledge integration framework that can automatically update the knowledge base with new research findings, (2) enhancing the reasoning capabilities through advanced model architectures and training strategies, and (3) improving the system's practical utility through better validation mechanisms and more efficient model designs. These improvements will make the system more robust, up-to-date, and accessible to the broader research community.

\section*{Acknowledgments}
This work was supported by the Advanced Materials-National Science and Technology Major Project (Grant No. 2025ZD0620100), Guangdong Basic and Applied Basic Research Foundation  2025A1515010304, Guangzhou Science and Technology Planning Project 2025A03J4491
\bibliography{anthology,acl2021}

\clearpage

\appendix


\section{Schema in Perovskite-KG}
\label{appendix:schema_in_Perovskite_KG}

Table \ref{tab:schema_kg} presents a comprehensive schema for the Perovskite-KG, organized into three main ontological categories: Fabrication, Parameters, and Performance. The Fabrication ontology encompasses process-related attributes such as coating parameters, methods, and annealing conditions. The Parameters ontology covers structural and compositional aspects including solvents, device architecture, and additives. The Performance ontology captures various stability metrics and efficiency parameters like thermal stability, light stability, and power conversion efficiency. Each category is further detailed with specific data types and examples to ensure precise knowledge representation. This structured schema enables systematic organization and retrieval of perovskite solar cell information while maintaining data consistency across the knowledge graph.

\begin{table*}[htbp]
  \centering
  \scalebox{0.6}{
\begin{tabular}{@{}c|cccc@{}}
\toprule
Ontology &
  Sub-Category &
  Data Type &
  Description &
  Example \\ \midrule
\multirow{3}{*}{Fabrication} &
  \begin{tabular}[c]{@{}c@{}}Coating \\ Parameter\end{tabular} &
  Float &
  \begin{tabular}[c]{@{}c@{}}The specifics of the coating method used \\ in the material deposition process.\end{tabular} &
  5000 rpm, 100$\mu$l \\ \cmidrule(l){2-5} 
 &
  Method &
  String &
  \begin{tabular}[c]{@{}c@{}}Different fabrication techniques, \\ involving variations in material deposition.\end{tabular} &
  spin coating \\ \cmidrule(l){2-5} 
 &
  \begin{tabular}[c]{@{}c@{}}Annealing \\ Parameter\end{tabular} &
  Float &
  \begin{tabular}[c]{@{}c@{}}Refers to the heating conditions applied to the perovskite, \\ which are essential for crystallization and stability.\end{tabular} &
  120°C, 10min \\ \midrule
\multirow{3}{*}{Parameters} &
  Solvent &
  String &
  \begin{tabular}[c]{@{}c@{}}the liquid medium used to dissolve precursors, \\ helping to form a uniform perovskite layer\end{tabular} &
  Dimethylformamide (DMF) \\ \cmidrule(l){2-5} 
 &
  \begin{tabular}[c]{@{}c@{}}Device \\ Structure\end{tabular} &
  \begin{tabular}[c]{@{}c@{}}Patterned \\ String\end{tabular} &
  \begin{tabular}[c]{@{}c@{}}The architecture of the device \\ (e.g., layer order, material interfaces)\end{tabular} &
  \begin{tabular}[c]{@{}c@{}}ITO/SAM/perovskite\\ /C60/BCP/Cu\end{tabular} \\ \cmidrule(l){2-5} 
 &
  Additive &
  String &
  Any additional materials or chemicals &
  potassium ions  \\ \midrule
\multirow{7}{*}{Performance} &
  \begin{tabular}[c]{@{}c@{}}Thermal \\ Stability\end{tabular} &
  String &
  \begin{tabular}[c]{@{}c@{}}The material's ability to \\ withstand heat without degrading\end{tabular} &
  \begin{tabular}[c]{@{}c@{}}\textgreater{}98\% of initial efficiency of \textgreater{}24\% \\ after 1,500 hours of continuous \\ maximum power point tracking\end{tabular} \\ \cmidrule(l){2-5} 
 &
  \begin{tabular}[c]{@{}c@{}}Light \\ Stability\end{tabular} &
  String &
  \begin{tabular}[c]{@{}c@{}}How resistant the material is \\ to prolonged exposure to light.\end{tabular} &
  \begin{tabular}[c]{@{}c@{}}\textgreater{}92\% of initial performance for 1,200 hours \\ under the damp-heat test \\ (85°C and 85\% relative humidity)\end{tabular} \\ \cmidrule(l){2-5} 
 &
  \begin{tabular}[c]{@{}c@{}}Moisture \\ Stability\end{tabular} &
  String &
  \begin{tabular}[c]{@{}c@{}}The material's resilience against \\ humidity or water exposure.\end{tabular} &
  \begin{tabular}[c]{@{}c@{}}Initial PCE of control, target-1 and target-2 \\ devices is 21.73\%, 24.42\% and 24.11\%, respectively. \\ Degraded to 78\% of initial PCE after 1,500 hours at 55±5°C\end{tabular} \\ \cmidrule(l){2-5} 
 &
  \begin{tabular}[c]{@{}c@{}}Fill Factor\\ Value\end{tabular} &
  Float &
  A measure of the device's maximum power output. &
  0.88 \\ \cmidrule(l){2-5} 
 &
  \begin{tabular}[c]{@{}c@{}}Open-Circuit \\ Voltage Value\end{tabular} &
  Float &
  \begin{tabular}[c]{@{}c@{}}The maximum voltage the device can \\ produce under open-circuit conditions.\end{tabular} &
  1.2 V \\ \cmidrule(l){2-5} 
 &
  \begin{tabular}[c]{@{}c@{}}Short-Circuit \\ Current Value\end{tabular} &
  Float &
  The current density when the circuit is closed. &
  25 mA/cm$^2$ \\ \cmidrule(l){2-5} 
 &
  \begin{tabular}[c]{@{}c@{}}Power Conversion \\ Efficiency Value\end{tabular} &
  Float &
  \begin{tabular}[c]{@{}c@{}}The efficiency with which the device \\ converts sunlight into electricity.\end{tabular} &
  25 \% \\ \bottomrule
\end{tabular}
}
\caption{Schema in Perovskite-KG.}
  \label{tab:schema_kg}
\end{table*}

\section{Prompts}
The system employs four specialized agents, each with carefully designed prompts to perform specific tasks in the perovskite solar cell knowledge processing pipeline:

1. \textbf{Information Extraction Agent} (Table \ref{tab:QuestionPrompts_agent_1}): Processes research papers using a structured set of 20 predefined questions across seven key categories, including device structure, performance enhancement, stability, and materials. The agent returns answers in a standardized JSON format, marking unavailable information as "Not mentioned" to maintain data quality.

2. \textbf{Verification Agent} (Table \ref{tab:VerificationPrompts_agent}): Validates extracted information by comparing it with source texts, focusing on maintaining accuracy of technical details like numerical values and material names. The agent provides both corrected content and justification for any modifications made.

3. \textbf{Organization Agent} (Table \ref{tab:OrganizationPrompts_agent}): Synthesizes verified information from multiple sources into coherent, topic-focused responses. This agent ensures that complex technical information is presented in a logical and accessible manner.

4. \textbf{LLM-Judge} (Table \ref{tab:LLM-JudgePrompts}): Evaluates response quality across four key criteria: accuracy, completeness, relevance, and clarity. Using a 1-5 scoring system, this agent provides detailed assessments and explanations for each criterion, along with an overall evaluation summary.

For LLM using in Perovskite-KG and Perovskite-Chat, we mainly using GPT-4o and GPT-4o-Mini for extraction and verification.
\begin{table*}[ht]
  \centering
  \fontsize{9}{11}\selectfont
  \caption{Prompts for Information Extraction Agent.}
  \begin{tabular}{p{0.97\textwidth}}
  \toprule
  \footnotesize \textbf{Prompts for Information Extraction Agent:} \\
  \midrule
  Answer the following questions based on the provided text. \\
  \{ \\
  \quad "Device Structure and Fabrication": [ \\
  \quad\quad "Q1: Summarize the device structures or configurations of the perovskite solar cells those can reach PCE over 25\%.", \\
  \quad\quad "Q2: How to prepare the perovskite precursor solutions those can reach PCE over 25\%?", \\
  \quad\quad "Q3: How to fabricate the perovskite solar cells those can reach PCE over 25\%?" \\
  \quad ], \\
  \quad "Performance Enhancement Strategies": [ \\
  \quad\quad "Q4: What are problems solved in literatures that report perovskite solar cells those can reach PCE over 25\%?", \\
  \quad\quad "Q5: What are the reasons to choose the strategies that can enhance performance of the perovskite solar cells in literatures that report perovskite solar cells those can reach PCE over 25\%?" \\
  \quad ], \\
  \quad "Performance Metrics Improvement": [ \\
  \quad\quad "Q6: How to improve the VOC of perovskite solar cells?", \\
  \quad\quad "Q7: How to improve the FF of perovskite solar cells?", \\
  \quad\quad "Q8: How to improve the Jsc of perovskite solar cells?" \\
  \quad ], \\
  \quad "Stability Improvements": [ \\
  \quad\quad "Q9: How to improve the moisture stability of perovskite solar cells?", \\
  \quad\quad "Q10: How to improve the thermal stability of perovskite solar cells?", \\
  \quad\quad "Q11: How to improve the illumination or light stability of perovskite solar cells?" \\
  \quad ], \\
  \quad "Defect and Recombination Management": [ \\
  \quad\quad "Q12: How to passivate or reduce defects/traps of perovskite solar cells?", \\
  \quad\quad "Q13: How to reduce recombination of perovskite solar cells?" \\
  \quad ], \\
  \quad "Interface and Extraction Layer Enhancements": [ \\
  \quad\quad "Q14: How to improve the wettability of the buried interface in perovskite solar cells?", \\
  \quad\quad "Q15: How to improve the hole extraction ability of HTL in perovskite solar cells?", \\
  \quad\quad "Q16: How to improve the electron extraction ability of ETL in perovskite solar cells?" \\
  \quad ], \\
  \quad "Materials Used in Perovskite Solar Cells": [ \\
  \quad\quad "Q17: What are the HTL materials used in perovskite solar cells and the common features of them?", \\
  \quad\quad "Q18: What are the ETL materials used in perovskite solar cells and their features?", \\
  \quad\quad "Q19: What are the hole blocking layer materials in perovskite solar cells and their features?", \\
  \quad\quad "Q20: What are the passivation materials used in perovskite solar cells and their common features?" \\
  \quad ] \\
  \} \\
  Below is the text: \texttt{\{paper\_text\}} \\
  Response: Return a JSON object with the following structure, if the text does not contain the answer, return "Not mentioned": \\
  \{ \\
  \quad "questions": [ \\
  \quad\quad \{ \\
  \quad\quad\quad "question": "Q1", \\
  \quad\quad\quad "answer": "Answer to Question 1" \\
  \quad\quad \}, \\
  \quad\quad \{ \\
  \quad\quad\quad "question": "Q2", \\
  \quad\quad\quad "answer": "Not mentioned" \\
  \quad\quad \}, \\
  \quad\quad ... \\
  \quad ] \\
  \} \\
  \bottomrule
  \end{tabular}
  \vspace{-2pt}
  \label{tab:QuestionPrompts_agent_1}
  \end{table*}

  \begin{table*}[ht]
    \centering
    \fontsize{9}{11}\selectfont
    \caption{Prompts for Verification Agent.}
    \begin{tabular}{p{0.97\textwidth}}
    \toprule
    \footnotesize \textbf{Prompts for Verification Agent:} \\
    \midrule
    You need to verify the accuracy of the extracted information from a perovskite paper. Compare the extracted data with the original text to ensure consistency and correctness. Highlight any discrepancies and fix them. Moreover, maintain the original meaning of the text and the original information, such as numbers and material names. \\
    \\
    Input: \\
    Paragraph \texttt{\{Section\_name\}}:\texttt{\{Text\_of\_the\_section\}} \\
    Extracted: \texttt{\{Extracted\_information\}} \\
    Output: Verified information with notes on any discrepancies or confirmation of accuracy. \\
    Please return a JSON object with the following structure only return one item: \\
    \{ \\
    \quad "verified\_info": \{ \\
    \quad\quad "fixed\_content": "The fixed paragraph", \\
    \quad\quad "reason": "The reason for the fix" \\
    \quad \} \\
    \} \\
    \bottomrule
    \end{tabular}
    \vspace{-2pt}
    \label{tab:VerificationPrompts_agent}
    \end{table*}

    \begin{table*}[ht]
      \centering
      \fontsize{9}{11}\selectfont
      \caption{Prompts for Organization Agent.}
      \begin{tabular}{p{0.97\textwidth}}
      \toprule
      \footnotesize \textbf{Prompts for Organization Agent:} \\
      \midrule
      Your task is to organize the verified information from a perovskite paper related to the question: \texttt{\{question\}}. \\
      Below is the information split into paragraphs that answers the question: \\
      \texttt{\{answers\}} \\
      \\
      Output: The organized and continuous answer to the question. \\
      \\
      Return a JSON object with the following structure: \\
      \{ \\
      \quad "answer": "The organized and continuous answer to the question." \\
      \} \\
      \bottomrule
      \end{tabular}
      \vspace{-2pt}
      \label{tab:OrganizationPrompts_agent}
      \end{table*}

  \begin{table*}[ht]
    \centering
    \fontsize{9}{11}\selectfont
    \caption{Prompts for LLM-Judge.}
    \begin{tabular}{p{0.97\textwidth}}
    \toprule
    \footnotesize \textbf{Prompts for LLM-Judge:} \\
    \midrule
    You are an expert evaluator. Your task is to compare a model's response to the ground truth answer and provide a detailed evaluation. \\
    \\
    Model's response: \\
    \texttt{\{model\_response\}} \\
    \\
    Ground truth: \\
    \texttt{\{ground\_truth\}} \\
    \\
    Please evaluate the model's response based on the following criteria: \\
    1. Accuracy: How factually correct is the model's response compared to the ground truth? \\
    2. Completeness: Does the model's response cover all the key points mentioned in the ground truth? \\
    3. Relevance: How well does the model's response address the implied question or task? \\
    4. Clarity: Is the model's response clear and easy to understand? \\
    \\
    For each criterion, provide a score from 1 to 5, where 1 is the lowest and 5 is the highest. Also, provide a brief explanation for each score. \\
    \\
    Finally, give an overall score from 1 to 5 and a summary of your evaluation. \\
    \\
    Format your response as a JSON object with the following structure: \\
    \{ \\
    \quad "accuracy": \{ "score": 0, "explanation": "" \}, \\
    \quad "completeness": \{ "score": 0, "explanation": "" \}, \\
    \quad "relevance": \{ "score": 0, "explanation": "" \}, \\
    \quad "clarity": \{ "score": 0, "explanation": "" \}, \\
    \quad "overall": \{ "score": 0, "summary": "" \} \\
    \} \\
    \bottomrule
    \end{tabular}
    \vspace{-2pt}
    \label{tab:LLM-JudgePrompts}
    \end{table*}
\section{Instruction Tuning Dataset}
\label{appendix:instruction_tuning_dataset}

\subsection{Dataset Statistics}
\label{appendix:dataset_statistics}
The research questions in perovskite solar cell studies are systematically categorized in Tables \ref{tab:question_classification} and \ref{tab:detailed_questions}. Table \ref{tab:question_classification} provides a high-level overview of seven major research categories, including Device Structure and Fabrication, Performance Enhancement Strategies, Performance Metrics Improvement, Stability Improvements, Defect and Recombination Management, Interface and Extraction Layer Enhancements, and Materials Used in Perovskite Solar Cells. Each category is accompanied by a detailed rationale explaining its scope and relevance. Table \ref{tab:detailed_questions} further expands this classification by presenting 21 specific research questions (Q1-Q21) organized within these seven categories. The questions cover a wide range of technical aspects, from device architecture optimization and performance enhancement strategies to material characteristics and stability improvements. Each research question is paired with its corresponding technical focus, providing a comprehensive framework for understanding the key areas of investigation in high-performance perovskite solar cell research.

\begin{table*}[htbp]
  \centering
  \setlength{\tabcolsep}{8pt}  
  \renewcommand{\arraystretch}{1.2}  

  \begin{tabular}{p{0.08\textwidth}p{0.47\textwidth}p{0.35\textwidth}}
    \toprule
    \textbf{ID} & \textbf{Research Question} & \textbf{Technical Focus} \\
    \midrule
    \multicolumn{3}{l}{\textit{\textbf{I. Device Structure and Fabrication}}} \\
    Q1 & Summarize device structures for PCE \textgreater 25\% & Device architecture optimization \\
    Q2 & Perovskite precursor solution preparation for PCE \textgreater 25\% & Solution chemistry and processing \\
    Q3 & Fabrication methods for PCE \textgreater 25\% & Manufacturing processes \\
    \midrule
    \multicolumn{3}{l}{\textit{\textbf{II. Performance Enhancement Strategies}}} \\
    Q4 & Problems solved in high-efficiency (\textgreater 25\%) devices & Critical challenges and solutions \\
    Q5 & Rationale for performance enhancement strategies & Strategic approach justification \\
    \midrule
    \multicolumn{3}{l}{\textit{\textbf{III. Performance Metrics Improvement}}} \\
    Q6 & V\textsubscript{OC} improvement methods & Open-circuit voltage optimization \\
    Q7 & FF improvement methods & Fill factor enhancement \\
    Q8 & J\textsubscript{SC} improvement methods & Short-circuit current density optimization \\
    Q9 & PLQY-iV\textsubscript{OC} relationship & Photoluminescence quantum yield correlation \\
    \midrule
    \multicolumn{3}{l}{\textit{\textbf{IV. Stability Improvements}}} \\
    Q10 & Moisture stability enhancement & Water resistance strategies \\
    Q11 & Thermal stability enhancement & Temperature tolerance methods \\
    Q12 & Light stability enhancement & Photo-stability improvement \\
    \midrule
    \multicolumn{3}{l}{\textit{\textbf{V. Defect and Recombination Management}}} \\
    Q13 & Defect/trap passivation methods & Defect control strategies \\
    Q14 & Recombination reduction approaches & Charge recombination suppression \\
    \midrule

    \multicolumn{3}{l}{\textit{\textbf{VI. Interface and Extraction Layer Enhancements}}} \\
    Q15 & Buried interface wettability improvement & Interface engineering \\
    Q16 & HTL hole extraction enhancement & Hole transport optimization \\
    Q17 & ETL electron extraction enhancement & Electron transport optimization \\
    \midrule
    \multicolumn{3}{l}{\textit{\textbf{VII. Materials Used in Perovskite Solar Cells}}} \\
    Q18 & HTL materials and features & Hole transport materials \\
    Q19 & ETL materials and features & Electron transport materials \\
    Q20 & Hole blocking layer materials & Blocking layer characteristics \\
    Q21 & Passivation materials and features & Surface passivation materials \\
    \bottomrule
  \end{tabular}
  \caption{Systematic Classification of Research Questions in High-Performance Perovskite Solar Cell Studies}
  \label{tab:detailed_questions}
\end{table*}

Table \ref{tab:category_names} shows the distribution of research categories in perovskite solar cells. Device Structure and Fabrication dominates the field, accounting for 24,198 entries (43.8\% of total). Performance Enhancement Strategies represents the second largest category with 11,233 entries (20.3\%), followed by Defect and Recombination Management with 7,209 entries (13.0\%). Stability Improvements, a crucial aspect of perovskite solar cell development, comprises 5,399 entries (9.8\%), while Performance Metrics Improvement accounts for 4,527 entries (8.2\%). Materials Used in Perovskite Solar Cells and Interface and Extraction Layer Enhancements represent smaller but significant portions of the research focus, with 1,586 (2.9\%) and 952 (1.7\%) entries respectively.

\begin{table*}[htbp]
  \centering
  \begin{tabular}{lll}
      \hline
      \textbf{Abbreviated Name} & \textbf{Full Name} & \textbf{Count} \\
      \hline
      Perf. Enhancement & Performance Enhancement Strategies & 11,233 \\
      Stability & Stability Improvements & 5,399 \\
      Defect \& Recom. & Defect and Recombination Management & 7,209 \\
      Device Structure & Device Structure and Fabrication & 24,198 \\
      Metrics & Performance Metrics Improvement & 4,527 \\
      Materials & Materials Used in Perovskite Solar Cells & 1,586 \\
      Interface & Interface and Extraction Layer Enhancements & 952 \\
      \hline
  \end{tabular}
  \caption{Correspondence between abbreviated and full names of research categories in perovskite solar cells}
  \label{tab:category_names}
\end{table*}

\begin{figure*}[t]
  \centering
  \includegraphics[width=1\linewidth]{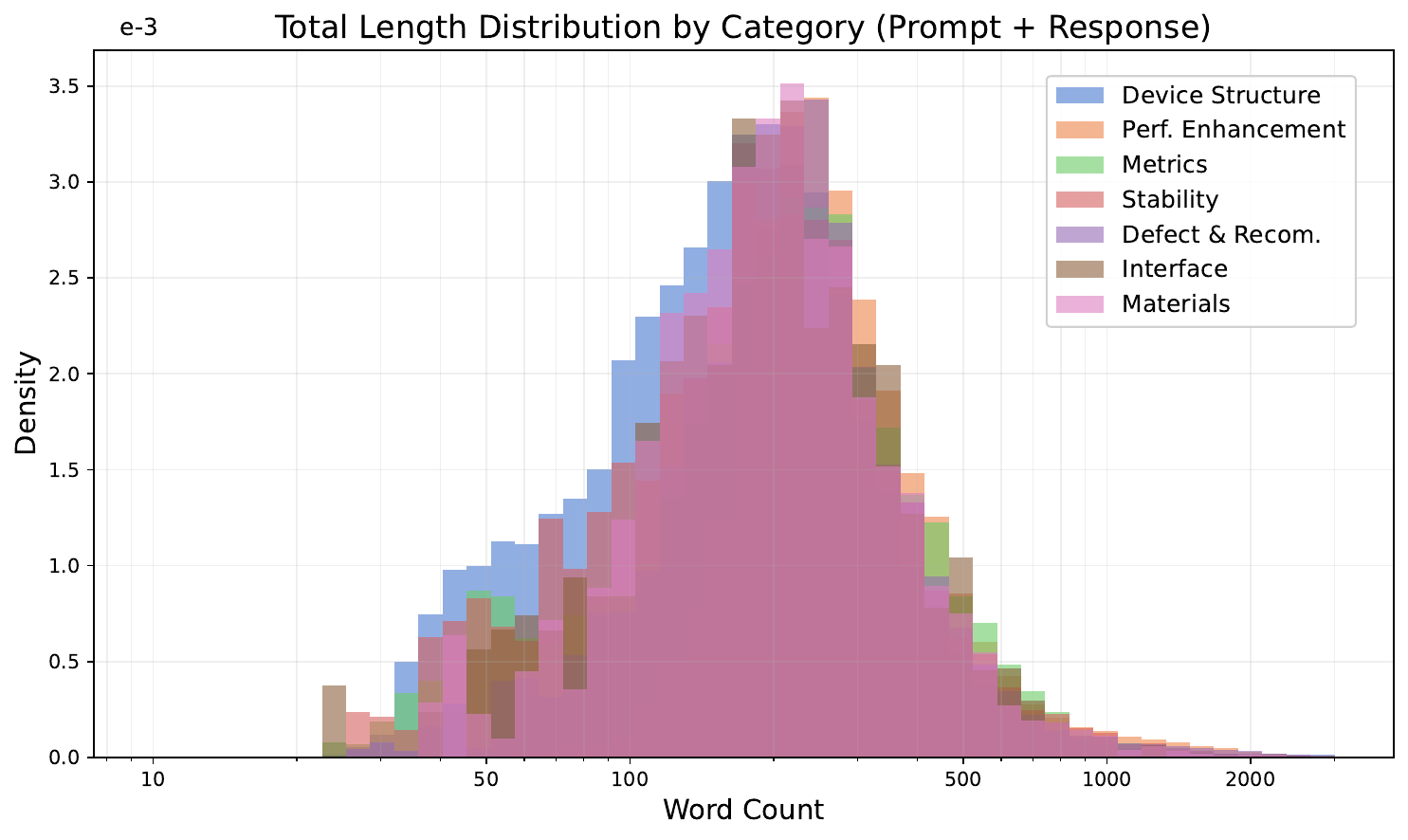}
  \caption{Distribution of prompt and response lengths across different categories in our dataset. The y-axis represents density (e-3), and the x-axis shows the word count in logarithmic scale. Each category's distribution is independently normalized.}
  \vspace{-0.4cm}
  \label{fig:total_length_distribution}
\end{figure*}

Figure \ref{fig:total_length_distribution} shows the length distribution patterns across categories in our perovskite instruction tuning dataset. All categories display similar characteristics, with sequence lengths predominantly ranging from 100 to 500 tokens and a median length of 400 tokens. This analysis informs our model design decisions, particularly regarding sequence length limitations and tokenization strategies. The distributions are independently normalized for each category, enabling direct pattern comparison despite varying sample sizes.

\begin{figure*}
  \centering
  \includegraphics[width=1\linewidth]{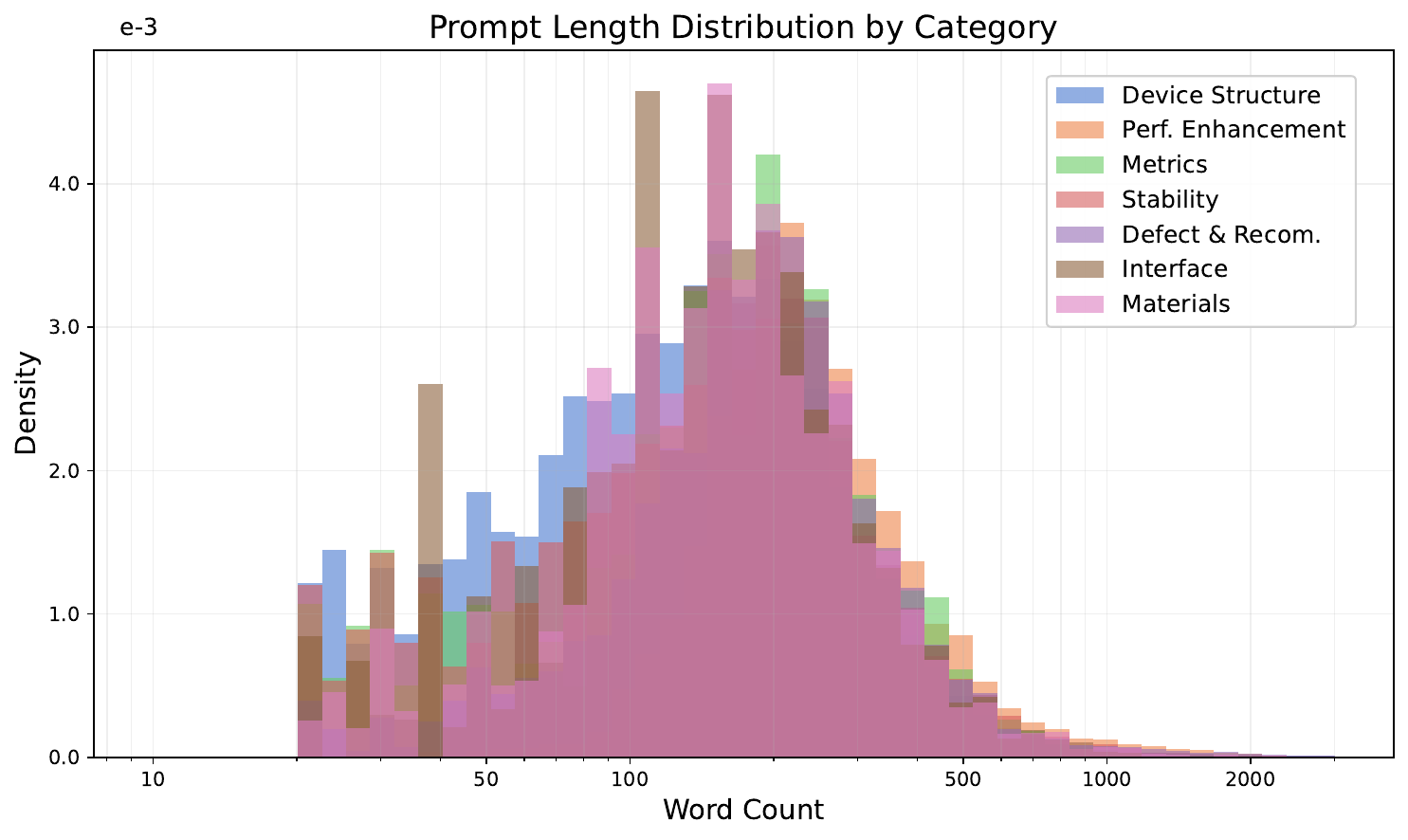}
  \includegraphics[width=1\linewidth]{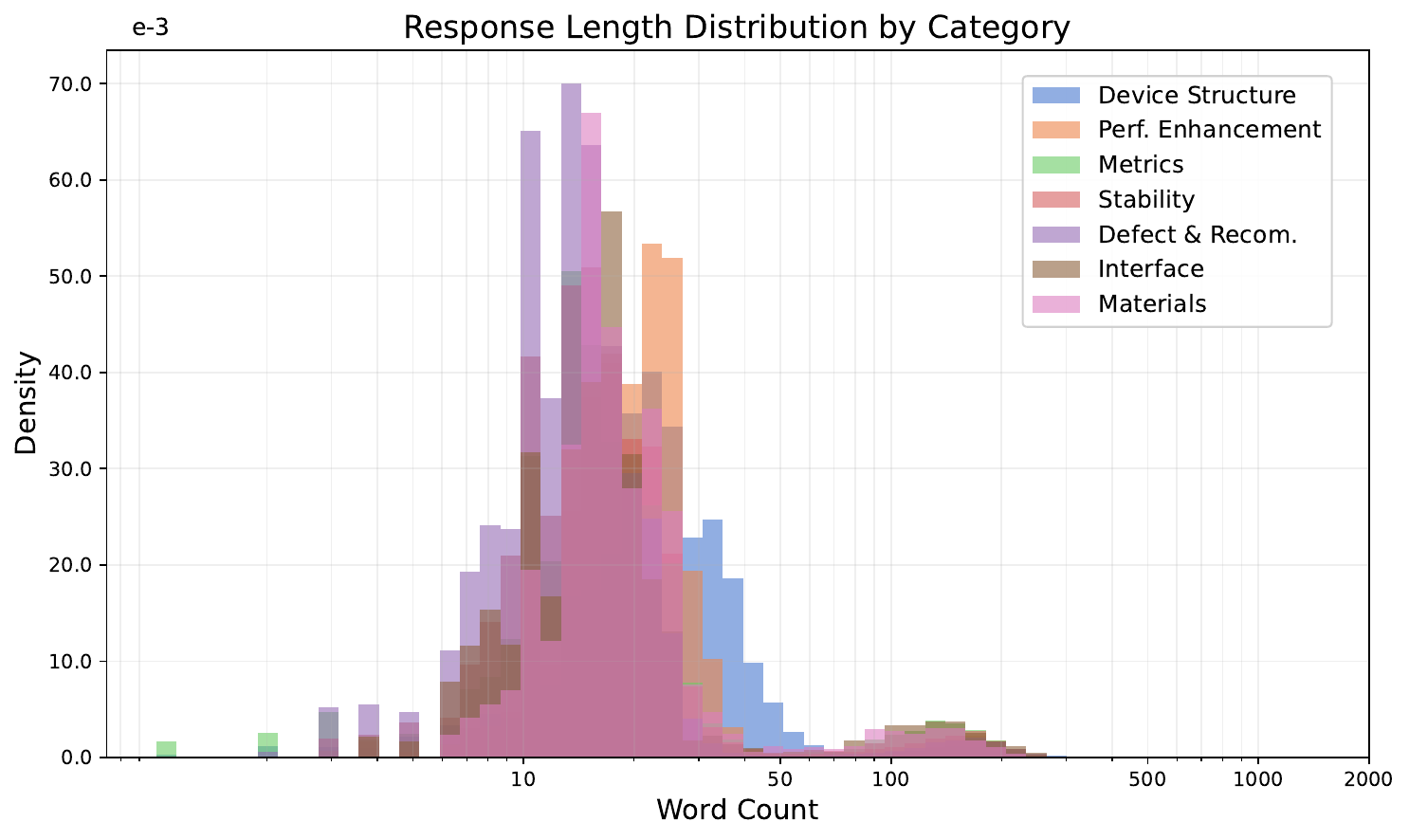}

  \caption{Distribution of prompt and response lengths across different categories in our dataset (top: prompt, bottom: response).}
  \label{fig:prompt_response_length_distribution}
\end{figure*}

Figure \ref{fig:prompt_response_length_distribution} illustrates the length distribution patterns across different categories in our perovskite instruction tuning dataset. 

\section{Instruction Tuning}
\label{app:tuning}
Table \ref{tab:training_params} presents the training hyperparameters for Perovskite-LLM.
\label{appendix:instruction_tuning}
\begin{table*}[h]
  \centering
  \begin{tabular}{l|cc}
    \toprule
    \textbf{Parameter} & \textbf{Chat-LLM} & \textbf{Reasoning-LLM} \\
    \midrule
    Learning Rate & 1e-6 & 5e-6\\
    Batch Size & 8 & 32 \\
    Number of Epochs & 3 & 15\\
    Optimizer & AdamW & AdamW\\
    Learning Rate Scheduler & Cosine & Cosine\\
    Warmup Steps & 20 & 20\\
    \bottomrule
  \end{tabular}
  \caption{Training Hyperparameters for Perovskite-Chat-LLM and Perovskite-Reasoning-LLM}

  \label{tab:training_params}
\end{table*}

\section{More Case Study}
\label{appendix:more_case_study}

\begin{figure*}[!t]
  \centering
  \tikzstyle{every node}=[font=\small,scale=0.9]
  \begin{tikzpicture}
      \footnotesize
      \definecolor{chatcolor1}{HTML}{FFFACD} 
      \definecolor{shadecolor}{gray}{0.95}
      \fontfamily{cmss}\selectfont
  
      \node[align=left, text width=0.8\textwidth, fill=shadecolor, rounded corners=1mm, draw=black, line width=0.5mm, fill opacity=0.7, text opacity=1, anchor=north west] (node1) at (-3.5cm,0) {
          \textbf{Question:}\\
          How to fabricate the perovskite solar cells those can reach PCE over 25\%?
      };
      \node[align=left, text width=0.8\textwidth, fill=chatcolor1, rounded corners=1mm, draw=black, line width=0.5mm, fill opacity=0.7, text opacity=1, anchor=north east] (node2) at ($(node1.south -| {0.6\textwidth,0}) + (0cm,-0.2cm)$) {
          \textbf{Perovskite-Chat-LLM:} \\
          To fabricate perovskite solar cells that can achieve a power conversion efficiency (PCE) over 25\%, several key steps and considerations are essential. First, the fabrication process should start with a clean and dry substrate, typically glass or flexible substrates, which are cleaned thoroughly using a combination of detergents and solvents. The substrates are then treated with UV ozone to enhance their wetting properties.

          For the hole transport layer (HTL), materials like PEDOT:PSS are commonly used. These materials are spin-coated onto the substrate and annealed at a temperature of \textbf{150°C} for \textbf{15 minutes}. The perovskite precursor solution is then prepared, typically using a mixed cation and halide composition, and spin-coated onto the HTL. The spin-coating process involves a two-step procedure: first at \textbf{1,000 rpm} for \textbf{10 seconds}, followed by \textbf{6,000 rpm} for \textbf{30 seconds}, with anisole added during the second step to facilitate crystallization.
          
          After spin-coating, the perovskite films are annealed at \textbf{100°C} for \textbf{10 minutes}. The electron transport layer (ETL) is then deposited, often using a solution of PC61BM in chlorobenzene, which is spin-coated at \textbf{1,000 rpm} for \textbf{30 seconds} and annealed at \textbf{100°C} for \textbf{10 minutes}. Finally, a metal electrode, typically gold, is deposited through thermal evaporation. The entire process is conducted in a nitrogen-filled glovebox to minimize exposure to moisture.
          
          };      
      
        \node[align=left, text width=0.8\textwidth, fill=chatcolor1, rounded corners=1mm, draw=black, line width=0.5mm, fill opacity=0.7, text opacity=1, anchor=north east] (node2) at ($(node2.south -| {0.6\textwidth,0}) + (0cm,-0.2cm)$) {
          \textbf{ChatGPT:} \\
          To fabricate perovskite solar cells with a power conversion efficiency (PCE) over 25\%, several key factors need to be considered and optimized in the fabrication process. Here are some steps that can help in achieving high-efficiency perovskite solar cells:

          1. Perovskite Material Quality: The quality of the perovskite material is crucial for achieving high efficiency. Use high-quality perovskite precursors and optimize the fabrication process to ensure uniform and defect-free perovskite films.
          
          2. Optimal Perovskite Layer Thickness: The thickness of the perovskite layer plays a significant role in determining the efficiency of the solar cell. Optimize the thickness of the perovskite layer to maximize light absorption and charge carrier generation.
          
          3. Tuning the Bandgap: By engineering the composition of the perovskite material, you can tune the bandgap to match the solar spectrum for better light absorption and improved efficiency.
          
          4. Interface Engineering: Pay attention to the interfaces between different layers in the solar cell structure. Optimize the interfaces to minimize recombination losses and improve charge carrier transport.
          
          5. Passivation Layers: Use passivation layers to reduce defects and trap states in the perovskite layer, which can improve the overall performance of the solar cell.
          
          6. Electron and Hole Transport Layers: Choose suitable electron and hole transport materials to facilitate efficient charge extraction and transport within the solar cell.
          };      
        
  \end{tikzpicture}
  \caption{A full conversation between Perovskite-LLM and ChatGPT, Perovskite-LLM provides detailed operational steps with specific parameters, while ChatGPT only offers general conceptual guidance.}
  \label{fig:appendix_example_perovskite_llm_chatgpt_1}
  \end{figure*}

  \begin{figure*}[!t]
    \centering
    \tikzstyle{every node}=[font=\small,scale=0.9]
    \begin{tikzpicture}
        \footnotesize
        \definecolor{chatcolor1}{HTML}{FFFACD} 
        \definecolor{shadecolor}{gray}{0.95}
        \fontfamily{cmss}\selectfont
    
        \node[align=left, text width=0.8\textwidth, fill=shadecolor, rounded corners=1mm, draw=black, line width=0.5mm, fill opacity=0.7, text opacity=1, anchor=north west] (node1) at (-3.5cm,0) {
            \textbf{Question:}\\
            What are the ETL materials used in perovskite solar cells and the features of them?
        };
        \node[align=left, text width=0.8\textwidth, fill=chatcolor1, rounded corners=1mm, draw=black, line width=0.5mm, fill opacity=0.7, text opacity=1, anchor=north east] (node2) at ($(node1.south -| {0.6\textwidth,0}) + (0cm,-0.2cm)$) {
            \textbf{Perovskite-Chat-LLM:} \\
            In perovskite solar cells, various electron transport layer (ETL) materials are utilized, each with distinct features. Common ETL materials include:

            1. Tin Oxide (SnO2): SnO2 is a popular choice for ETLs due to its high electron mobility, low-temperature process ability, and compatibility with flexible substrates. It is often used in n-i-p structured perovskite solar cells.
            
            2. Zinc Oxide (ZnO): ZnO is another widely used ETL material, known for its high electron mobility and low-temperature processing. It is commonly used in both n-i-p and p-i-n structured perovskite solar cells.
            
            3. Tungsten Oxide (WO3): WO3 is utilized as an interlayer between the perovskite layer and the ETL, enhancing charge extraction and improving device stability. It is particularly effective in p-i-n structured devices.
            
            4. Phenyl-C61-Butyric Acid Methyl Ester (PCBM): PCBM is a fullerene derivative used as an ETL in p-i-n structured perovskite solar cells. It offers high electron mobility and is compatible with various perovskite compositions.
            
            5. Tin Oxide Doped with Fluorine (F-SnO2): This material is used to enhance the performance of SnO2 ETLs, improving their electron mobility and stability, which leads to higher efficiency in perovskite solar cells.
            
            6. Cesium Lead Halide (CsPbX3): CsPbX3 is employed as a wide-bandgap ETL in all-inorganic perovskite solar cells, contributing to improved stability and efficiency. \\
            ...
            };      
        
          \node[align=left, text width=0.8\textwidth, fill=chatcolor1, rounded corners=1mm, draw=black, line width=0.5mm, fill opacity=0.7, text opacity=1, anchor=north east] (node2) at ($(node2.south -| {0.6\textwidth,0}) + (0cm,-0.2cm)$) {
            \textbf{ChatGPT:} \\
            ETL (Electron Transport Layer) materials play a crucial role in perovskite solar cells by facilitating the efficient extraction and transport of electrons generated by the perovskite layer. Some commonly used ETL materials in perovskite solar cells include:

            1. TiO2 (Titanium Dioxide): TiO2 is one of the most widely used ETL materials in perovskite solar cells. It has good electron transport properties, high electron mobility, and is chemically stable. TiO2 can be easily processed into thin films, making it suitable for large-scale production of solar cells.
            
            2. SnO2 (Tin Oxide): SnO2 is another popular ETL material in perovskite solar cells. It has a high electron mobility, good transparency, and can be deposited at low temperatures. SnO2 films can also be easily doped to improve their conductivity and electron extraction efficiency. \\
            ...

            };      
          
    \end{tikzpicture}
    \caption{A full conversation between Perovskite-LLM and ChatGPT, Perovskite-LLM provides detailed information about the ETL materials and their features, while ChatGPT only offers general information about the ETL materials.}
    \label{fig:appendix_example_perovskite_llm_chatgpt_2}
    \end{figure*}

    \begin{figure*}[!t]
      \centering
      \tikzstyle{every node}=[font=\small,scale=0.9]
      \begin{tikzpicture}
          \footnotesize
          \definecolor{chatcolor1}{HTML}{FFFACD} 
          \definecolor{shadecolor}{gray}{0.95}
          \fontfamily{cmss}\selectfont
      
          \node[align=left, text width=0.8\textwidth, fill=shadecolor, rounded corners=1mm, draw=black, line width=0.5mm, fill opacity=0.7, text opacity=1, anchor=north west] (node1) at (-3.5cm,0) {
              \textbf{Question:}\\
              How to improve the thermal stability of perovskite solar cells?
          };
          \node[align=left, text width=0.8\textwidth, fill=chatcolor1, rounded corners=1mm, draw=black, line width=0.5mm, fill opacity=0.7, text opacity=1, anchor=north east] (node2) at ($(node1.south -| {0.6\textwidth,0}) + (0cm,-0.2cm)$) {
              \textbf{Perovskite-Chat-LLM:} \\
              Improving the thermal stability of perovskite solar cells involves several strategies. One effective approach is the use of thermally stable materials for the charge transporting layer, such as nickel oxide (NiO) and copper thiocyanate (CuSCN), which can enhance the thermal stability of the devices. Additionally, the incorporation of additives like 1,8-diodoctane (DIO) and 1-chloronaphthalene (CN) has been shown to improve the morphology of perovskite films, leading to better thermal stability. Furthermore, the use of thermally stable hole transport materials (HTMs) and electron transport materials (ETMs) is crucial, as they can withstand high temperatures without degrading the perovskite layer. The development of new HTMs, such as poly(3-hexylthiophene) (P3HT), has also contributed to improved thermal stability. Lastly, the introduction of interfacial layers, like graphene oxide, can enhance the thermal stability of the perovskite layer by reducing defects and improving charge transport. \\
              ...

              };      
          
            \node[align=left, text width=0.8\textwidth, fill=chatcolor1, rounded corners=1mm, draw=black, line width=0.5mm, fill opacity=0.7, text opacity=1, anchor=north east] (node2) at ($(node2.south -| {0.6\textwidth,0}) + (0cm,-0.2cm)$) {
              \textbf{ChatGPT:} \\
              Improving the thermal stability of perovskite solar cells is crucial for their long-term performance and reliability. Here are some strategies to enhance their thermal stability:

              1. Encapsulation: One of the most effective ways to improve the thermal stability of perovskite solar cells is to encapsulate them with protective materials. Encapsulation helps to shield the cells from environmental factors such as moisture, oxygen, and heat, which can degrade the perovskite material.
              
              2. Improved Perovskite Composition: Researchers are constantly working on developing new perovskite compositions that are more stable at high temperatures. By tweaking the chemical composition of the perovskite material, it is possible to enhance its thermal stability. \\
              ...
              };      
            
      \end{tikzpicture}
      \caption{A full conversation between Perovskite-LLM and ChatGPT, Perovskite-LLM provides detailed information about the thermal stability of perovskite solar cells, while ChatGPT only offers general information about the thermal stability of perovskite solar cells.}
      \label{fig:appendix_example_perovskite_llm_chatgpt_3}
      \end{figure*}

      \begin{figure*}[!t]
        \centering
        \tikzstyle{every node}=[font=\small,scale=0.9]
        \begin{tikzpicture}
            \footnotesize
            \definecolor{chatcolor1}{HTML}{FFFACD} 
            \definecolor{shadecolor}{gray}{0.95}
            \fontfamily{cmss}\selectfont
        
            \node[align=left, text width=0.8\textwidth, fill=shadecolor, rounded corners=1mm, draw=black, line width=0.5mm, fill opacity=0.7, text opacity=1, anchor=north west] (node1) at (-3.5cm,0) {
                \textbf{Question:}\\
                Can you tell me what is CuO? And give me 5 usage in perovskite.
            };
            \node[align=left, text width=0.8\textwidth, fill=chatcolor1, rounded corners=1mm, draw=black, line width=0.5mm, fill opacity=0.7, text opacity=1, anchor=north east] (node2) at ($(node1.south -| {0.6\textwidth,0}) + (0cm,-0.2cm)$) {
                \textbf{Perovskite-Chat-LLM:} \\
                CuO refers to copper(II) oxide, a chemical compound that plays a significant role in various applications, particularly in perovskite solar cells (PSCs). Here are five key usages of CuO in perovskite technology:
                
                1. \textbf{Hole Transport Material}: CuO acts as an effective hole transport material (HTM) due to its suitable work function and high hole mobility. It is often used in conjunction with other materials, such as $\bm{NiO_x}$, to enhance the performance of PSCs by facilitating efficient charge extraction [1].
                
                2. \textbf{Interface Optimization}: The use of CuO helps in optimizing the interfaces within perovskite solar cells. By depositing CuO films on top of perovskite layers, it can improve the interfacial properties and reduce recombination losses at these critical junctions[2]. 
                
                3. \textbf{Thermal Stability Enhancement}: Copper oxide (CuO) has been shown to provide thermal stability to perovskite materials, which is crucial for maintaining high efficiency during prolonged operation under elevated temperatures. This property makes it an attractive choice for applications where environmental stability is a concern [3].
                
                4. \textbf{Doping and Passivation}: CuO can be utilized as a dopant in other materials used in PSCs, such as SnO2, enhancing their electrical properties like conductivity and mobility. Additionally, the doping effects can lead to improved passivation of defects at the perovskite layer, contributing to better performance metrics [4].
                
                5. \textbf{Integration with Other Materials}: In bifacial solar cells, CuO has been integrated with materials such as carbon nanotubes (CNTs) to create high-performance architectures that benefit from synergistic effects. This integration not only enhances charge transport but also improves the overall efficiency of the devices [5]. \\
                
                \textbf{References:} \\
    
                [1] Inorganic Materials as Hole Selective Contacts and Intermediate Tunnel Junction Layer for Monolithic Perovskite‐CIGSe Tandem Solar Cells, Wang, Yajie; Wenisch, Robert; Schlatman \\ 
    
                [2] Critical Review of Cu‐Based Hole Transport Materials for Perovskite Solar Cells: From Theoretical Insights to Experimental Validation, Sun, Qingde; Sadhu, Anupam; Lie, Stener; Wong, Lydia Helena, Advanced Materials, 2024 \\
    
                [3] Processing‐Performance Evolution of Perovskite Solar Cells: From Large Grain Polycrystalline Films to Single Crystals, Haque, Md Azimul; Troughton, Joel; Baran, Derya, Advanced Energy Materials, 2020 \\
    
                [4] Air‐Processed Perovskite Solar Cells with >25\% Efficiency and High Stability Enabled by Crystallization Modulation and Holistic Passivation, Shi, Xiaoyu; Liu, Tianxiao; Dou, Yunjie; Hu, Xiaodong; Liu, Yangyang; Wang, Feifei; Wang, Lingyuan; Ren, Zhijun; Chen, Shangshang, Advanced Materials, 2024 \\
                
                [5] Critical Review of Cu‐Based Hole Transport Materials for Perovskite Solar Cells: From Theoretical Insights to Experimental Validation, Sun, Qingde; Sadhu, Anupam; Lie, Stener; Wong, Lydia Helena, Advanced Materials, 2024
                };      
            
              
        \end{tikzpicture}
        \caption{A case study of Perovskite-Chat-LLM's ability to provide detailed and accurate information with references.}
        \label{fig:example_perovskite_llm_cite}
        \end{figure*}

\section{License}
GPQA~\citep{rein2023gpqa} and Minerva~\citep{lewkowycz2022solving} are under MIT license.
\end{document}